\documentclass{article}

\usepackage{float}
\usepackage[margin=1.0in]{geometry}
\usepackage{abstract}
\usepackage[utf8]{inputenc}

\usepackage[dvipsnames,svgnames,table,xcdraw]{xcolor}
\usepackage{relsize}
\usepackage{tcolorbox}
\usepackage{listings}
\usepackage{subcaption}
\usepackage{comment}
\usepackage{graphicx}
\usepackage{booktabs}
\usepackage{multirow}
\usepackage{todonotes}
\usepackage{enumitem}
\usepackage{url}
\usepackage{mathpazo}
\usepackage{amsmath,amssymb,amsthm,amsfonts,bm}
\usepackage[toc,page,header]{appendix}
\usepackage{fancyhdr}
\usepackage[
  backend=bibtex,
  natbib=true,
  maxbibnames=99,
  minalphanames=3
]{biblatex}
\usepackage[colorlinks=true]{hyperref}
\usepackage[T1]{fontenc}
\usepackage{newunicodechar}
\usepackage{auth_detailed}

\usepackage{mathtools}
\usepackage{longtable,tabularx}
\usepackage{placeins}
\usepackage{threeparttable}
\usepackage{adjustbox}
\usepackage{pifont}

\counterwithout{figure}{section}
\counterwithout{table}{section}

\definecolor{brown}{HTML}{4E3629}
\definecolor{plotYellow}{HTML}{FFB000}
\definecolor{plotOrange}{HTML}{FE6100}
\definecolor{plotMagenta}{HTML}{DC267F}
\definecolor{plotPurple}{HTML}{785EF0}
\definecolor{plotBlue}{HTML}{648FFF}

\hypersetup{
  colorlinks=true,
  urlcolor=plotBlue,
  citecolor=plotBlue,
  linkcolor=plotBlue
}

\usepackage{sectsty}
\allsectionsfont{\color{brown}}

\usepackage{array}
\usepackage{makecell}
\usepackage{wrapfig}
\usepackage{needspace}
\usepackage{afterpage}
\usepackage{pgfplots}
\usetikzlibrary{arrows.meta,calc}
\usepgfplotslibrary{groupplots}
\pgfplotsset{compat=1.18}
\pgfplotscreateplotcyclelist{paperPalette}{plotBlue,plotMagenta,plotOrange,plotPurple,plotYellow}
\pgfplotsset{every axis/.append style={cycle list name=paperPalette}}

\newunicodechar{–}{--}
\newunicodechar{—}{---}
\newunicodechar{‘}{`}
\newunicodechar{’}{'}
\newunicodechar{“}{``}
\newunicodechar{”}{''}

\DeclareMathOperator*{\argmax}{arg\,max}

\tcbuselibrary{skins, breakable}

\usepackage{parskip}

\colorlet{oursShade}{plotOrange!12}
\colorlet{oursShadeLight}{white}
\colorlet{rowShade}{brown!5}
\newcommand{\best}[1]{\textbf{#1}}
\newcommand{\second}[1]{\underline{#1}}

\usepackage{fontawesome5}
\newcommand{\researchquestion}[1]{%
  \begin{center}
  {\Large\color{plotYellow}\faLightbulb[regular]}\hspace{0.6em}%
  {\large\bfseries\itshape\color{brown}#1}
  \end{center}}

\input{template_adapter}
\newcommand{\runningtitle}{SwarmEvolve}

\begin{document}

\title{SwarmEvolve: Latent Agent Swarms for Evolutionary Discovery and Optimization}

\author[1]{Qile Jiang}
\author[1]{George Em Karniadakis}
\affiliation[1]{Division of Applied Mathematics, Brown University}


\abstract{
LLM-guided evolutionary search uses AI agents to improve programs for mathematical and computational problems by iteratively refining solutions and combining high-scoring methods. However, the process is often slow because the LLM generates a long textual chain of thought for each round and must repeat this generation many times to maintain population diversity.
We present \textbf{SwarmEvolve}, a framework in which hundreds of low-cost agents per round work in swarms, reason in latent space without decoding to text, and share information by passing KV caches. Each swarm studies the problem from a different angle. A novelty filter keeps the most distinct agent ideas by directly measuring the distances between the agents' hidden state vectors. Crossover of ideas happens naturally in latent space, as a summarizer agent receives the combined KV caches of all swarm members. 
On 21 problems in mathematical discovery and scientific computing, SwarmEvolve gets the highest average score and needs a median of only 3 to 8 minutes to reach the best scores that other state-of-the-art frameworks reach in 60 minutes. On several problems, SwarmEvolve with Qwen3 8B improves on the solutions that AlphaEvolve first found with frontier LLMs. 
Our work shows that agent swarms that communicate in latent space offer a fast, low-cost approach to program search for scientific problems.

\vspace{6pt}{\centering\href{https://github.com/Qile-J/SwarmEvolve}{\color{brown}\faGithub\ \texttt{github.com/Qile-J/SwarmEvolve}}\par}}

\maketitle


\begin{figure}[H]
\centering
\includegraphics[width=\textwidth]{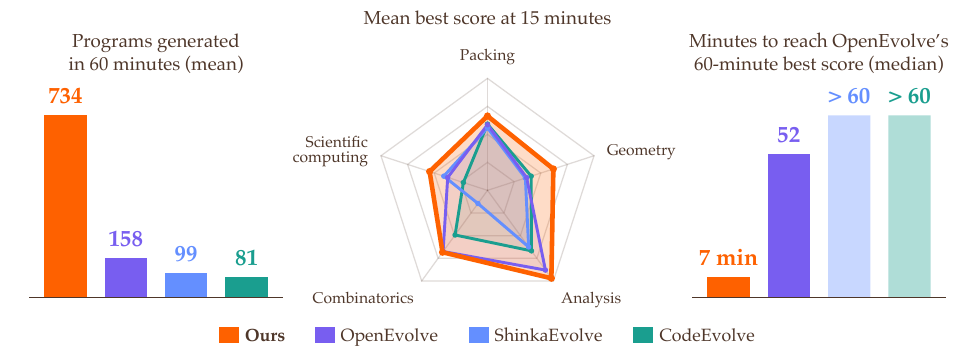}
\caption{SwarmEvolve (ours) compared with OpenEvolve, ShinkaEvolve, and CodeEvolve on 21 problems, with Qwen3 8B on one GPU for 60 minutes per run. Left: total number of programs generated. Middle: mean best score at 15 minutes. Right: median minutes to reach OpenEvolve's 60-minute score. SwarmEvolve has a higher throughput and better scores.}
\label{fig:headline}
\end{figure}

\section{Introduction}

Many problems in mathematics and scientific computing can be framed as a search for a good program. Examples include symbolic regression, optimizing constants in inequalities, and building robust solvers for differential equations. This search can be carried out through evolutionary computation, a method to iteratively improve candidate programs through variation, crossover, and selection~\citep{holland1975adaptation,koza1992genetic}. 

Recently, LLMs have been used as the mutation and crossover operators in such evolutionary processes~\citep{lehman2024elm,meyerson2024lmx,wu2025roadmap}. These LLM-guided frameworks have made strong progress on many open problems. In particular, FunSearch found new cap set constructions~\citep{romeraparedes2024funsearch}, and AlphaEvolve found a more efficient algorithm for multiplying $4 \times 4$ complex matrices~\citep{novikov2025alphaevolve}. Later work applied this approach to solve many more problems in mathematics and computer systems~\citep{georgiev2026scale,dupont2026matrix,cheng2025barbarians}. These frameworks differ mainly in how they explore. FunSearch evolves programs on separate islands, and AlphaEvolve and its open-source variants add a MAP-Elites archive to keep diverse programs~\citep{sharma2025openevolve,assumpcao2025codeevolve,khrulkov2025gigaevo}. To save LLM calls, some methods reject near-duplicate proposals, adaptively allocate the search budget, or evolve the search strategy itself~\citep{lange2025shinkaevolve,cemri2026adaevolve,liu2026evox}. Others guide exploration with better context, such as how past changes affected the score, related literature, simulated peer review, or a solver split into separate steps~\citep{jiang2026deltaevolve,liu2025deepevolve,yu2025alpharesearch,kravatskiy2026improvevolve}.

Recent work also scales up the search. Some methods run many parallel trajectories~\citep{ye2026simpletes} or train the LLM on each problem with reinforcement learning~\citep{yuksekgonul2026tttdiscover,surina2025evotune,wang2026thetaevolve}. Others use more agents with different roles, such as an ensemble of frontier LLMs in AlphaEvolve, multi-agent debate, and coding agents that coordinate through shared memory or an orchestrator~\citep{jiang2025agenticsciml,qu2026coral,virk2026swarmresearch}. When agents build on one another’s findings, they have produced new results on open problems~\citep{cai2026cogentic,bianchi2026collective}. These works suggest that diverse proposals, cross-pollination of ideas, and iterative refinement are key to new discoveries on open scientific problems.

Yet in most of these frameworks, the search is slow and expensive. Each agent writes a long chain of thought in text. The growing solution archive, evaluator feedback, and text messages between agents make the LLM context very long. Scaling to many agents, larger models, or test-time training multiplies this cost. As a result, the search cannot run at the scale that many open problems need. We therefore ask:

\researchquestion{Can swarms of low-cost agents make evolutionary search faster and better?}

We present \textbf{SwarmEvolve}, an evolutionary search framework in which swarms of low-cost latent agents reason in latent space and communicate by passing KV caches. It builds on recent work on latent reasoning, where an LLM reasons with continuous vectors instead of sampled tokens~\citep{hao2025coconut,zhang2025softthinking,shen2025codi,xu2025softcot,zhu2025superposition}, and on latent communication, where LLMs exchange token embeddings~\citep{pham2024cipher}, activations~\citep{ramesh2025activations}, hidden states~\citep{zheng2025thoughtcomm,du2026interlat,zou2026recursivemas}, or KV caches~\citep{shi2026kvcomm,fu2026c2c} instead of text messages. In particular, LatentMAS combines both without training. Its agents reason in latent space and pass their KV caches to a summarizer that writes the answer~\citep{zou2026latentmas}. We use this mechanism as the basis for large-scale, low-cost evolutionary search.

Because latent reasoning and communication are cheap, SwarmEvolve runs hundreds of agents in each round of evolution. The agents form distinct swarms, and each swarm has its own focus, such as implementing a construction from the literature or using a textbook method. Inside a swarm, each agent reasons from its unique perspective for 40 latent steps. A novelty filter then keeps the agents in each swarm whose latent reasoning trajectories are farthest apart in cosine distance. Finally, each swarm’s summarizer agent reads the KV caches of the kept agents and writes one new program.

Compared with existing frameworks, SwarmEvolve has three advantages:

\begin{enumerate}
\item \textbf{Greater diversity.} While existing methods generate one response at a time or run a small team of agents, SwarmEvolve runs 450 latent agents per round, each with a different specialization, parent program, and way of thinking, and so samples far more diverse proposals.
\item \textbf{Built-in crossover.} SwarmEvolve recombines ideas directly in latent space. A summarizer agent reads the KV caches of agents that studied different parent programs and writes one program that combines their ideas.
\item \textbf{Higher throughput.} Since agents do not decode or encode text between steps, SwarmEvolve evaluates 5 to 13 times more programs per hour than existing SOTA frameworks and reaches their 60-minute scores in a median of 3 to 8 minutes.
\end{enumerate}

We compare SwarmEvolve with OpenEvolve, ShinkaEvolve, and CodeEvolve on 21 problems from mathematical discovery and scientific computing. Each framework runs with Qwen3 8B, 14B, and 32B-FP8~\citep{yang2025qwen3} for 60 minutes on one GPU (\autoref{fig:headline}). SwarmEvolve gets the highest average score with all three models. With Qwen3 8B, it scores higher than all three baselines with Qwen3 32B-FP8. After only 15 minutes, it already scores higher than the 60-minute averages of all baselines with every model. On three problems, SwarmEvolve with a frozen Qwen3 8B also improves on the constructions that AlphaEvolve originally found with frontier Gemini models. These results suggest that swarms of low-cost latent agents are a fast and scalable way to use AI for open mathematical and scientific discovery.

\section{Preliminaries}

In this section, we introduce the preliminary setup and notations for our method. 

\subsection{LLMs and the KV cache}

We use decoder-only transformer LLMs~\citep{vaswani2017attention}, which embed the input token at position $t$ as a vector $\mathbf{e}_t \in \mathbb{R}^d$, where $d$ is the hidden dimension, and pass the sequence through $L$ transformer layers. The output of the last layer at position $t$ is the last hidden state $\mathbf{h}_t \in \mathbb{R}^d$. The LLM computes the distribution of the next token from $\mathbf{h}_t$.

Inside each layer, the query $\mathbf{q}_t$ of position $t$ attends to the keys $\mathbf{k}_j$ and values $\mathbf{v}_j$ of all positions $j \le t$. Because the keys and values of a position depend only on the tokens up to that position, the LLM stores them in a KV cache $\{(\mathbf{K}^{(\ell)}, \mathbf{V}^{(\ell)})\}_{\ell=1}^{L}$, where $\mathbf{K}^{(\ell)}$ and $\mathbf{V}^{(\ell)}$ stack the keys and values of all positions in layer $\ell$. Each new token then costs one forward pass of a single position, which attends to the KV cache and appends its own keys and values to it. The KV cache of a shared prefix, such as a problem statement, is usually computed once and reused by every sequence that continues it for efficiency.

The Qwen3 models that we use encode positions with rotary position embeddings (RoPE)~\citep{su2024roformer,yang2025qwen3}. RoPE multiplies each query and key at position $i$ by a rotation matrix $\mathbf{R}_i$ whose angles grow linearly with $i$, so the attention score depends only on the offset between the two positions:
\begin{align}
(\mathbf{R}_i \mathbf{q})^\top (\mathbf{R}_j \mathbf{k}) &= \mathbf{q}^\top \mathbf{R}_{j-i}\, \mathbf{k}.
\label{eq:rope}
\end{align}
Moving a block of the KV cache from positions $0, 1, 2, \dots$ to positions $\delta, \delta+1, \delta+2, \dots$ therefore requires multiplying each cached key by $\mathbf{R}_\delta$, since the values carry no positional information.

\subsection{Latent reasoning}

LLMs usually reason by decoding a chain of thought in text~\citep{wei2022cot,kojima2022zeroshot}. In latent reasoning, the LLM instead feeds the last hidden state $\mathbf{h}_t$ back as the next input embedding $\mathbf{e}_{t+1}$, so it reasons in its continuous embedding space without sampling tokens~\citep{hao2025coconut,zhang2025softthinking,zhu2025survey}.

LatentMAS applies latent reasoning to LLM agents without training~\citep{zou2026latentmas}. After an agent reads its input, each latent reasoning step maps the last hidden state $\mathbf{h}_t$ directly to the next input embedding $\mathbf{e}_{t+1}$. This map has two parts. First, an alignment matrix $\mathbf{W}_{\mathrm{a}}$ maps $\mathbf{h}_t$ into the input embedding space. Second, the result is rescaled to norm $\gamma$, where $\gamma$ is the mean norm of the input token embeddings over the vocabulary. The matrix $\mathbf{W}_{\mathrm{a}}$ is the ridge regression solution, with regularization weight $\rho$, that maps the output embedding of each token to its input embedding:
\begin{align}
\mathbf{e}_{t+1} &= \gamma\, \frac{\mathbf{W}_{\mathrm{a}} \mathbf{h}_t}{\lVert \mathbf{W}_{\mathrm{a}} \mathbf{h}_t \rVert}, \quad \quad
\mathbf{W}_{\mathrm{a}} = \mathbf{W}_{\mathrm{in}}^\top \mathbf{W}_{\mathrm{out}} \big(\mathbf{W}_{\mathrm{out}}^\top \mathbf{W}_{\mathrm{out}} + \rho\, \mathbf{I}\big)^{-1},
\label{eq:latent_step}
\end{align}
where $\mathbf{W}_{\mathrm{in}}, \mathbf{W}_{\mathrm{out}}\in \mathbb{R}^{|\mathcal{V}| \times d}$ are respectively the input and output embedding matrices and $|\mathcal{V}|$ is the vocabulary size. The agent takes a fixed number of latent reasoning steps given by \autoref{eq:latent_step} and decodes no text. Each latent reasoning step adds one position to the KV cache.

\subsection{Latent communication}

LatentMAS agents communicate by passing KV caches instead of text~\citep{zou2026latentmas}. In the standard hierarchical setting, $J$ agents, usually with different specializations, each take their own latent reasoning steps. A summarizer agent then writes the final answer. Let $\mathbf{K}^{(\ell)}_j$ and $\mathbf{V}^{(\ell)}_j$ be the keys and values of agent $j$ in layer $\ell$, for $j = 1, \ldots, J$. Let $\mathbf{K}^{(\ell)}_{\mathrm{sum}}$ and $\mathbf{V}^{(\ell)}_{\mathrm{sum}}$ be the keys and values of the summarizer's own prompt in layer $\ell$. The KV cache of the summarizer concatenates them along the sequence dimension:
\begin{align}
\mathbf{K}^{(\ell)} &= [\mathbf{K}^{(\ell)}_1; \dots; \mathbf{K}^{(\ell)}_J; \mathbf{K}^{(\ell)}_{\mathrm{sum}}],  \quad \quad
\mathbf{V}^{(\ell)} = [\mathbf{V}^{(\ell)}_1; \dots; \mathbf{V}^{(\ell)}_J; \mathbf{V}^{(\ell)}_{\mathrm{sum}}] \quad \quad  \text{for each layer $\ell$}.
\label{eq:latentmas}
\end{align}
The summarizer then decodes the answer as text, attending to both its own inputs and the latent reasoning steps of all $J$ agents. When the agents run in parallel, the KV cache of each agent starts at position $0$. We therefore use the rotation in \autoref{eq:rope} to move each cache to its offset in the concatenation.


\begin{figure}[!t]
\centering
\includegraphics[width=\textwidth]{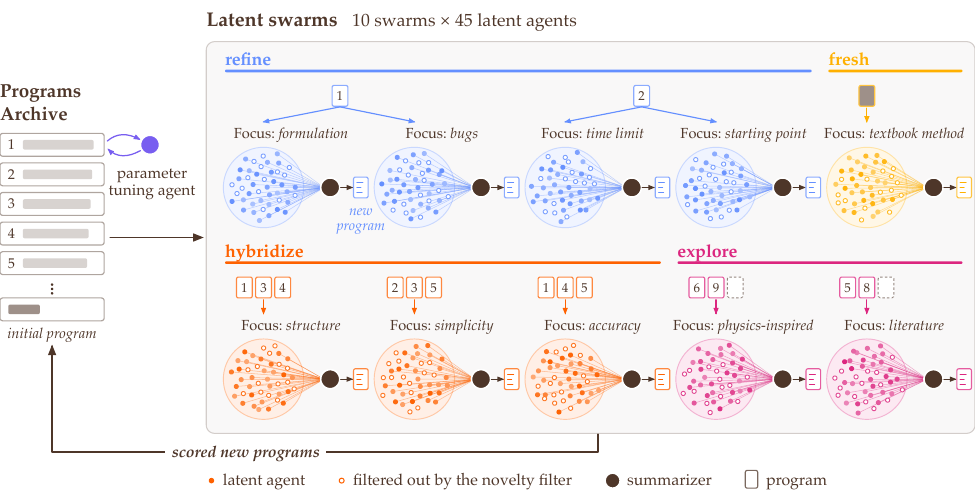}
\caption{Workflow of SwarmEvolve. In each round, SwarmEvolve launches 10 swarms of 45 latent agents. Each swarm draws parent programs from the archive and gets a focus prompt to refine, hybridize, explore, or start fresh. In each swarm, the agents take 40 latent reasoning steps, a novelty filter removes the 15 most similar agents (hollow circles), and the summarizer reads the KV caches of the rest to write one new program. The new programs are evaluated and added to the archive. A parameter tuning agent tunes the hyperparameters of the best program.}
\label{fig:method-overview}

\vspace{10pt}
\begin{minipage}[c]{0.585\textwidth}
\includegraphics[width=\linewidth]{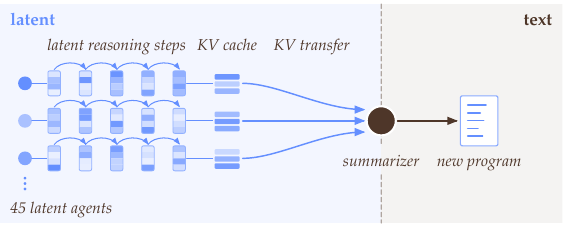}
\end{minipage}\hfill
\begin{minipage}[c]{0.37\textwidth}
\caption{Latent reasoning and communication in a swarm. Each latent agent reasons recursively in the embedding space without decoding to text. Its latent reasoning steps are stored as KV cache. The summarizer then reads the concatenated KV caches from all agents and writes a new program in text.}
\label{fig:latent-agents}
\end{minipage}
\end{figure}

\section{SwarmEvolve}
\label{sec:method}

SwarmEvolve is an evolutionary search framework that runs latent-reasoning agent swarms to improve programs over many rounds. It keeps an archive of all candidate programs, together with their scores and evaluation feedback. In each round, it selects parent programs from the archive for refinement and crossover. The search requires an initial program, which is usually a simple baseline solution. As illustrated in \autoref{fig:method-overview}, each round consists of the following steps.

\paragraph{Form swarms.} To start a round, SwarmEvolve forms 10 swarms of 45 latent agents each. All agents in a swarm share one main focus. Each agent receives a distinct prompt, which combines one of five phrasings of this focus with one of nine ways of thinking, such as reasoning from first principles or reconsidering assumptions (Supplementary Table~\ref{tab:focuses}). Each swarm also receives a set of parent programs, which are assigned to its agents in turn so that each agent reads exactly one program. The 10 swarms are of four types:

\begin{itemize}
\item \textbf{Four refinement swarms} each receive one of the two best programs in the archive, with two swarms per program. Each swarm takes the main focus used least often so far for its program, such as its mathematical formulation or its use of the time budget. Their goal is to improve the best programs.
\item \textbf{Three hybridization swarms} each receive five programs drawn at random from the ten best, so nine agents read each program. Their goal is to combine the strongest ideas into one new program.
\item \textbf{Two exploration swarms} each receive four programs drawn at random from the entire archive. Nine agents read each program, and the other nine agents read only the problem statement with no parent program. Their goal is to explore less considered directions.
\item \textbf{One fresh-start swarm} receives only the initial program. Its goal is to find a new approach that differs from existing ones.
\end{itemize}

\paragraph{Reason in latent space.} Each latent agent reads the problem statement, its focus prompt, and its assigned parent program with the program's evaluation results. It then takes 40 latent reasoning steps. All 450 agents run in one batch. Agents that read the same program attend to one shared copy of its KV cache through paged attention~\citep{kwon2023vllm}.

\paragraph{Filter for novelty.} In each swarm, the novelty filter keeps the 30 latent agents whose reasoning differs the most and discards the other 15. Let $\mathbf{h}_{j,t}$ be the last hidden state of agent $j$ at latent reasoning step $t$. Each agent is represented by its mean hidden state over the 40 steps,
\begin{align}
\bar{\mathbf{h}}_j &= \frac{1}{40} \sum_{t=1}^{40} \mathbf{h}_{j,t}, \quad \quad 
j = 1, \ldots, 45.
\label{eq:mean-hidden-state}
\end{align}
Then, each agent's mean hidden state $\bar{\mathbf{h}}_j$ is centered on the swarm mean $\boldsymbol{\mu}$ and normalized to a unit vector $\mathbf{z}_j = (\bar{\mathbf{h}}_j - \boldsymbol{\mu}) / \lVert \bar{\mathbf{h}}_j - \boldsymbol{\mu} \rVert$. The kept set $\mathcal{K}$ starts with the agent closest to $\boldsymbol{\mu}$. Greedy farthest-point selection~\citep{gonzalez1985clustering} in cosine distance then adds one agent at a time, selecting
\begin{align}
j^\star &= \argmax_{j \notin \mathcal{K}} \min_{i \in \mathcal{K}} \left( 1 - \mathbf{z}_j^\top \mathbf{z}_i \right),
\label{eq:novelty-filter}
\end{align}
until $\mathcal{K}$ holds 30 agents.

\paragraph{Transfer KV caches.} Each summarizer reads the reasoning of its 30 kept agents through one KV cache (\autoref{fig:latent-agents}). In each layer $\ell$, this KV cache is the concatenation
\begin{align}
\begin{aligned}
\mathbf{K}^{(\ell)} &= [\mathbf{K}^{(\ell)}_{\mathrm{prob}}; \mathbf{K}^{(\ell)}_{1}; \dots; \mathbf{K}^{(\ell)}_{30}; \mathbf{K}^{(\ell)}_{p_1}; \dots; \mathbf{K}^{(\ell)}_{p_Q}; \mathbf{K}^{(\ell)}_{\mathrm{ins}}], \\
\mathbf{V}^{(\ell)} &= [\mathbf{V}^{(\ell)}_{\mathrm{prob}}; \mathbf{V}^{(\ell)}_{1}; \dots; \mathbf{V}^{(\ell)}_{30}; \mathbf{V}^{(\ell)}_{p_1}; \dots; \mathbf{V}^{(\ell)}_{p_Q}; \mathbf{V}^{(\ell)}_{\mathrm{ins}}],
\end{aligned}
\label{eq:summarizer-cache}
\end{align}
where $\mathbf{K}^{(\ell)}_{\mathrm{prob}}$ and $\mathbf{V}^{(\ell)}_{\mathrm{prob}}$ are the keys and values of the problem statement. For $j = 1, \dots, 30$, $\mathbf{K}^{(\ell)}_{j}$ and $\mathbf{V}^{(\ell)}_{j}$ are the keys and values of the focus prompt and 40 latent reasoning steps of the $j$-th kept agent. For $q = 1, \dots, Q$, $\mathbf{K}^{(\ell)}_{p_q}$ and $\mathbf{V}^{(\ell)}_{p_q}$ are the keys and values of the $q$-th parent program $p_q$ of the swarm. $\mathbf{K}^{(\ell)}_{\mathrm{ins}}$ and $\mathbf{V}^{(\ell)}_{\mathrm{ins}}$ are the keys and values of the summarizer instruction. Before the concatenation, each latent agent's key $\mathbf{k}$ is rotated to $\mathbf{R}_{\delta}\mathbf{k}$ (\autoref{eq:rope}), where $\delta$ is the position of its block in $\mathbf{K}^{(\ell)}$. The values carry no positional information and are concatenated unchanged.

\paragraph{Write programs.} After receiving the KV cache, the summarizer synthesizes and decodes the latent reasoning traces of its agents. The summarizer outputs a short description of its direction and one complete program required to solve the problem.

\paragraph{Tune parameters.} Every round, a parameter tuning agent reads the best program, its evaluation results, and a list of all numeric parameters of the program. It then proposes a single new value for one of the parameters to improve the program. SwarmEvolve evaluates up to 8 such one-number edits per round.

\paragraph{Evaluate.} After new programs are generated, the framework discards programs that duplicate an earlier one, checks the syntax, and automatically adds missing imports of standard package names. The deterministic evaluator then scores each new program on a CPU core. When the evaluator finishes, the scored programs enter the archive with their scores, logs, and error messages.

SwarmEvolve iterates these steps for the time budget of the run. The best-scoring program in the archive at the end of the run is returned as the final solution.


\section{Experiments}
\label{sec:experiments}

\begin{table}[t]
\centering
\caption{Best score within 60 minutes, averaged over three seeds (scores defined in Section~\ref{sec:experiments} and Appendix~\ref{app:problems}). For each model, \textbf{bold} marks the best score and \underline{underline} marks the second best. SwarmEvolve (\textbf{ours}) has the highest average score with every model.}
\label{tab:results-60min}
\small
\setlength{\tabcolsep}{3pt}
\renewcommand{\arraystretch}{1.15}
\begin{adjustbox}{max width=\textwidth}
\begin{tabular}{ll>{\columncolor{oursShade}}cccc>{\columncolor{oursShade}}cccc>{\columncolor{oursShade}}cccc}
\toprule
& & \multicolumn{4}{c}{\color{brown}\textbf{Qwen3 8B @ 60 min}} & \multicolumn{4}{c}{\color{brown}\textbf{Qwen3 14B @ 60 min}} & \multicolumn{4}{c}{\color{brown}\textbf{Qwen3 32B-FP8 @ 60 min}} \\[-3pt]
& & \multicolumn{4}{c}{\makebox[0pt]{\footnotesize\color{black!60}1$\times$ L40S GPU, 8 CPU cores}} & \multicolumn{4}{c}{\makebox[0pt]{\footnotesize\color{black!60}1$\times$ RTX PRO 6000 GPU, 8 CPU cores}} & \multicolumn{4}{c}{\makebox[0pt]{\footnotesize\color{black!60}1$\times$ H100 GPU, 8 CPU cores}} \\
\cmidrule(lr){3-6}\cmidrule(lr){7-10}\cmidrule(lr){11-14}
\color{brown}\textbf{Group} & \color{brown}\textbf{Problem} & \textbf{Ours} & \makecell[cb]{Open-\\Evolve} & \makecell[cb]{Shinka-\\Evolve} & \makecell[cb]{Code-\\Evolve} & \textbf{Ours} & \makecell[cb]{Open-\\Evolve} & \makecell[cb]{Shinka-\\Evolve} & \makecell[cb]{Code-\\Evolve} & \textbf{Ours} & \makecell[cb]{Open-\\Evolve} & \makecell[cb]{Shinka-\\Evolve} & \makecell[cb]{Code-\\Evolve} \\
\midrule
\multirow{4}{*}{Packing} & Circles in a square & \best{0.998} & 0.877 & 0.876 & \second{0.990} & \best{0.999} & 0.868 & \second{0.996} & 0.991 & \best{0.997} & 0.915 & 0.987 & \second{0.992} \\
  & Circles in a rectangle & \cellcolor{oursShadeLight}0.888 & \second{0.969} & 0.958 & \best{0.987} & \cellcolor{oursShadeLight}\second{0.976} & 0.956 & 0.954 & \best{0.992} & \best{1.000} & 0.960 & 0.954 & \second{0.984} \\
  & Hexagons, $n=11$ & \best{0.627} & \second{0.546} & 0.491 & 0.491 & \best{0.771} & \second{0.688} & 0.545 & 0.540 & \best{0.874} & \second{0.780} & 0.697 & 0.491 \\
  & Hexagons, $n=12$ & \best{0.707} & \second{0.593} & 0.546 & 0.493 & \best{0.829} & \second{0.721} & 0.648 & 0.595 & \cellcolor{oursShadeLight}\second{0.650} & \best{0.786} & 0.521 & 0.493 \\
\midrule
\multirow{5}{*}{Geometry} & Heilbronn, triangle & \best{0.558} & 0.050 & 0.418 & \second{0.514} & \best{0.710} & 0.152 & 0.172 & \second{0.591} & \best{0.831} & 0.058 & 0.473 & \second{0.513} \\
  & Heilbronn, convex, $n=13$ & \best{0.724} & 0.601 & \second{0.656} & 0.610 & \best{0.739} & 0.617 & \second{0.660} & 0.641 & \best{0.878} & 0.570 & \second{0.704} & 0.679 \\
  & Heilbronn, convex, $n=14$ & \best{0.807} & 0.421 & 0.342 & \second{0.487} & \best{0.769} & 0.638 & \second{0.659} & 0.555 & \best{0.780} & 0.654 & \second{0.692} & 0.638 \\
  & Distance ratio, plane & \best{0.994} & 0.716 & 0.876 & \second{0.887} & \best{1.000} & 0.716 & \second{0.907} & 0.842 & \best{0.999} & 0.716 & \second{0.848} & 0.735 \\
  & Distance ratio, space & \best{0.994} & 0.817 & 0.890 & \second{0.913} & \best{0.999} & 0.880 & 0.903 & \second{0.924} & \best{0.999} & 0.897 & \second{0.915} & 0.887 \\
\midrule
\multirow{4}{*}{Analysis} & Autocorrelation I & \best{0.995} & \second{0.993} & 0.992 & 0.992 & \best{0.994} & \second{0.993} & \second{0.993} & 0.992 & \best{0.995} & \second{0.993} & \second{0.993} & 0.992 \\
  & Autocorrelation II & \best{1.021} & \second{1.006} & 0.994 & 1.003 & \best{1.032} & \second{1.013} & 1.004 & 0.994 & \best{1.016} & 0.985 & \second{0.992} & 0.985 \\
  & Autocorrelation III & \best{1.005} & 0.779 & 0.979 & \second{0.991} & \best{1.005} & 0.988 & \second{0.998} & 0.988 & \best{1.005} & \second{0.998} & 0.997 & 0.991 \\
  & Uncertainty & \best{0.911} & \second{0.899} & 0.000 & 0.780 & \best{0.912} & \second{0.899} & 0.000 & 0.889 & \cellcolor{oursShadeLight}\second{0.906} & \best{0.911} & 0.000 & 0.895 \\
\midrule
\multirow{3}{*}{Combinatorics} & Erd\H{o}s overlap & \best{0.999} & \second{0.998} & 0.997 & 0.961 & \best{1.000} & \best{1.000} & \second{0.999} & 0.997 & \best{0.999} & \best{0.999} & \best{0.999} & \second{0.998} \\
  & Sums and differences & \best{0.957} & \second{0.939} & 0.000 & 0.923 & \best{0.958} & \second{0.929} & 0.299 & 0.914 & \best{0.964} & 0.932 & 0.580 & \second{0.945} \\
  & Kissing number, $d=11$ & \cellcolor{oursShadeLight}0.309 & \second{0.481} & \best{0.508} & 0.371 & \cellcolor{oursShadeLight}0.570 & \best{0.782} & \second{0.617} & 0.371 & \cellcolor{oursShadeLight}\second{0.709} & \best{0.753} & 0.615 & 0.319 \\
\midrule
\multirow{5}{*}{\makecell[l]{Scientific\\computing}} & Sparse linear systems & \best{0.933} & 0.648 & \second{0.695} & 0.420 & \cellcolor{oursShadeLight}\second{0.867} & \best{0.881} & 0.594 & 0.664 & \best{1.000} & 0.814 & \second{0.903} & 0.800 \\
  & Eigenvalues & \cellcolor{oursShadeLight}\second{0.700} & \best{0.723} & \best{0.723} & \best{0.723} & \best{0.723} & \best{0.723} & \best{0.723} & \best{0.723} & \cellcolor{oursShadeLight}0.776 & \best{0.849} & \second{0.804} & 0.729 \\
  & Burgers equation & \best{0.741} & 0.661 & \second{0.682} & 0.568 & \cellcolor{oursShadeLight}0.715 & \second{0.716} & \best{0.724} & 0.641 & \best{0.833} & 0.750 & \second{0.792} & 0.662 \\
  & Convection-diffusion & \best{0.494} & \second{0.479} & 0.420 & 0.269 & \cellcolor{oursShadeLight}0.502 & \best{0.542} & \second{0.514} & 0.287 & \best{0.520} & \second{0.500} & 0.447 & 0.277 \\
  & Helmholtz equation & \best{0.529} & 0.346 & \second{0.447} & 0.323 & \best{0.517} & \second{0.496} & 0.486 & 0.397 & \best{0.515} & 0.430 & \second{0.487} & 0.348 \\
\midrule
\multicolumn{2}{l}{Average} & \best{0.804} & 0.693 & 0.642 & \second{0.700} & \best{0.837} & \second{0.771} & 0.686 & 0.739 & \best{0.869} & \second{0.774} & 0.733 & 0.731 \\
\bottomrule
\end{tabular}
\end{adjustbox}
\end{table}

We compare SwarmEvolve with three state-of-the-art frameworks: OpenEvolve~\citep{sharma2025openevolve}, ShinkaEvolve~\citep{lange2025shinkaevolve}, and CodeEvolve~\citep{assumpcao2025codeevolve} on 21 problems from five domains (Appendix~\ref{app:problems}). Sixteen of these problems, in packing, geometry, analysis, and combinatorics, come from Appendix B of AlphaEvolve~\citep{novikov2025alphaevolve}. For these problems, we use the same problem statements, initial programs, and evaluators from OpenEvolve. The score of each problem is its objective value divided by that of the construction published by AlphaEvolve, so a score above 1 improves on AlphaEvolve. 

We also design five scientific computing problems on sparse linear systems, eigenvalues of non-normal matrices, and the Burgers, convection-diffusion, and Helmholtz equations, whose score is the number of correct digits relative to a high-fidelity ground truth, averaged over challenging test cases with various parameters or matrices. Problem details are given in Appendix~\ref{app:problems}. 

Each framework is given 60 minutes on one GPU with 8 CPU cores that evaluate 8 programs in parallel. The three baselines keep their default configurations, let the model reason in text before it answers, and serve it with vLLM~\citep{kwon2023vllm} for fast inference. Every framework runs three seeds on every problem with each of Qwen3 8B, 14B, and 32B-FP8. 

\begin{figure}[t]
\centering
\includegraphics[width=\textwidth]{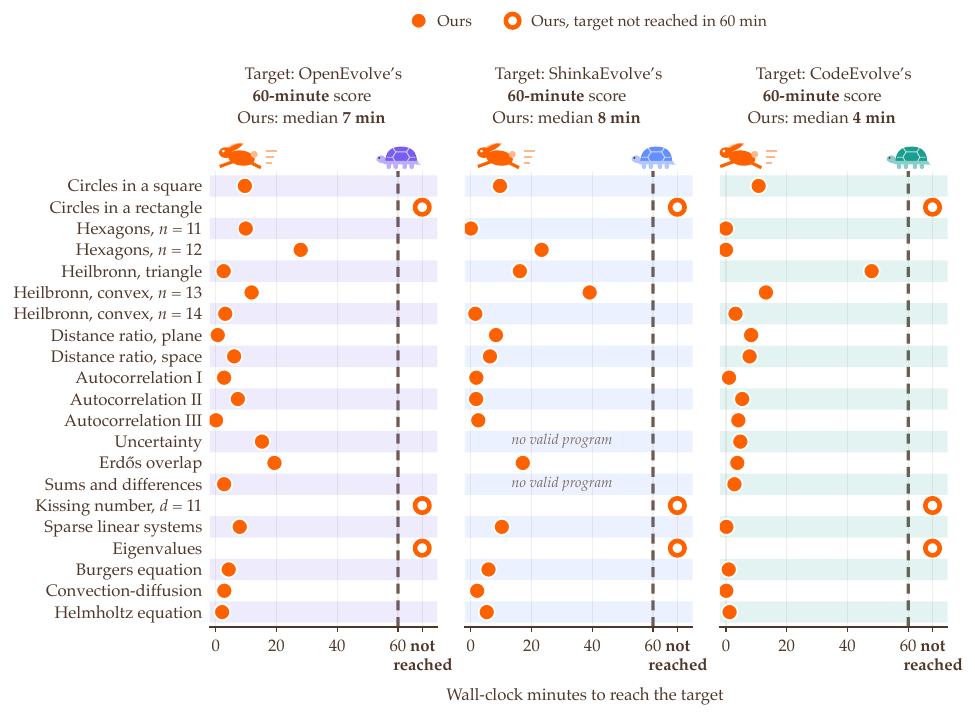}
\caption{Minutes that SwarmEvolve needs to reach the best score that OpenEvolve (left), ShinkaEvolve (middle), and CodeEvolve (right) reach in 60 minutes. All frameworks use Qwen3 8B and run for three seeds. Hollow circles mark problems where SwarmEvolve does not reach the target within 60 minutes. On most problems, SwarmEvolve reaches in a few minutes what the baselines reach in one hour.}
\label{fig:time-to-baseline}
\end{figure}

\subsection{Results}

As shown in \autoref{tab:results-60min}, SwarmEvolve achieves the highest average score with all three LLMs. It also achieves the best score on most individual problems. SwarmEvolve's average score with Qwen3 8B outperforms all three baselines running Qwen3 32B-FP8. With Qwen3 32B-FP8, SwarmEvolve solves all five sparse linear systems to full accuracy, which no baseline run ever achieves. In addition to higher scores, SwarmEvolve is also superior in the following ways: 

\paragraph{More programs per hour.}
SwarmEvolve evaluates 5 to 13 times more programs per hour than the baselines (\autoref{fig:headline}, left). This difference comes from how the frameworks spend GPU time. The baselines spend most of it on text reasoning, which takes 20 to 35 seconds per program. In contrast, the 450 latent agents of one round take only a few seconds in total. The novelty filter of SwarmEvolve also saves time, because it uses the hidden states of the agents directly. ShinkaEvolve instead needs an external embedding model and an LLM judge to reject similar proposals. Supplementary Table~\ref{tab:time-breakdown} gives a detailed time breakdown.

\paragraph{Faster to the same score.}
SwarmEvolve reaches the 60-minute scores of the baselines early in its run (\autoref{fig:time-to-baseline}). The median time is 4 to 8 minutes with Qwen3 8B and 3 to 6 minutes with the larger LLMs. Each round of SwarmEvolve takes about one minute and explores several parent programs at once. As a result, the first few rounds already cover many directions, so the best program is discovered sooner. 

\paragraph{Beyond AlphaEvolve with an 8B model.}
With Qwen3 8B, SwarmEvolve surpasses AlphaEvolve's published constructions on three problems, compared with one for each baseline (Supplementary Table~\ref{tab:alphaevolve-raw}). Its packing of 26 circles in a square reaches the best known sum of radii, 2.635983~\citep{specht2026packomania}, and its step functions improve AlphaEvolve's bounds for the second and third autocorrelation inequalities. While AlphaEvolve relies on frontier LLMs, SwarmEvolve achieves these results with swarms of small open-weight LLMs. 

Together, these results show that SwarmEvolve is a faster and more efficient framework for evolutionary search, which finds better programs than text-based frameworks even with a small open-weight LLM.

\subsection{Latent space analysis}

Latent agents make the reasoning of a swarm directly measurable and fast to filter numerically. In this section, we analyze how the reasoning trajectories of the swarms are spread across latent space.

\begin{figure}[t]
\centering
\includegraphics[width=\textwidth]{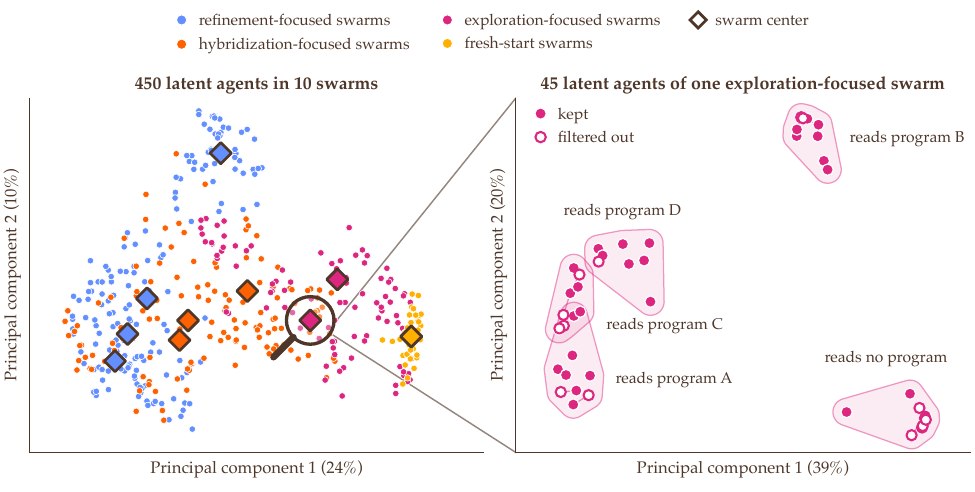}
\caption{Mean hidden states of the latent agents for a circle packing problem with Qwen3 8B, projected onto their first two principal components: the 450 latent agents of all 10 swarms with the swarm centers (left), and the 45 latent agents of one exploration-focused swarm, each reading different parent programs (right). The swarms occupy distinct regions, and the novelty filter keeps latent agents from every cluster.}
\label{fig:latent-space}
\end{figure}

\autoref{fig:latent-space} shows one round of a circle packing problem with Qwen3 8B. It projects the centered and normalized mean hidden states of the latent agents onto their first two principal components. 

On the left panel, agents of the same swarm lie close together and apart from other swarms, so each focus clearly moves its swarm to its own region of latent space. In this example, the first principal component roughly orders the swarms by role, from the refinement swarms to the exploration and fresh-start swarms.

On the right panel, within a swarm, the agents form clusters by the parent program they read. The novelty filter keeps agents from every cluster and covers the swarm roughly evenly. The focus prompts and the novelty filter therefore give each round proposals that differ both across swarms and within each swarm, leading to diversity in the search process. 

\section{Discussion}

SwarmEvolve shows that evolutionary search can be fast, cheap, and strong at the same time. Latent agents skip the slow step of writing text, so one GPU can run hundreds of agents in each round. With the same model and the same time budget, SwarmEvolve evaluates many more programs than text-based frameworks, reaches their final scores early in the run, and ends with higher scores. Even with a small open-weight LLM, it improves on several constructions that AlphaEvolve found with frontier LLMs.

Latent reasoning is a key part of this gain. When we replace the latent reasoning steps with text reasoning, the scores drop (Appendix~\ref{app:text-reasoning}), even when the agents are given a higher text token budget than the latent token budget. Longer text makes each round slower, so the search evaluates fewer programs in the same time and ends with lower scores. Latent reasoning gives each agent enough room to think while keeping each round short.

Scaling the number of agents in a swarm also helps to improve the framework performance early in the search (Appendix~\ref{app:swarm-size}). Since a larger swarm studies each parent program from more perspectives, each round explores more of the design space initially. This early advantage matters most when each evaluation is expensive, such as a large simulation in science or engineering, where only a few programs can be evaluated. Later in the search, swarms of different sizes reach similar scores, while scaling to larger, more intelligent LLMs still raises the final score consistently.

More broadly, SwarmEvolve makes large-scale search for new programs affordable on a single GPU. This opens the method to many problems in mathematics and scientific computing. In mathematics, swarms can search for new constructions and proof strategies. In computing, they can improve algorithms for speed and accuracy. In scientific computing, they can write better PDE solvers, discover formulas through symbolic regression, and design more efficient numerical methods. Many of these problems have clear evaluators or desired outcomes, such as a higher accuracy or faster speed, which is what evolutionary search needs.

SwarmEvolve has three limitations. First, all latent agents and summarizers share one LLM, because a summarizer can read the KV caches of its agents directly only when they come from the same model. Learned translations between the KV caches of different models~\citep{fu2026c2c} could extend SwarmEvolve to swarms whose agents run different models. Second, our implementation requires LLMs with standard attention. Linear-attention, state-space, and hybrid models compress the past into a fixed-size state~\citep{katharopoulos2020linear,gu2024mamba,yang2025gateddeltanet}, so latent communication between them needs a way to merge the states of several agents. Third, the memory for the KV caches of a round grows linearly with the number of agents, the number of latent reasoning steps, and the number of layers, so the largest LLM we test needs a large GPU. Compressed latent messages~\citep{chen2026condenseflow} and rounds spread over several GPUs could let SwarmEvolve scale to larger models and longer runs.

\section{Summary}

We introduce SwarmEvolve, an evolutionary search framework in which swarms of low-cost agents reason in latent space and communicate by passing KV caches. This design lets one GPU run hundreds of agents in each round, so the search explores much more of the design space in the same time. On problems from mathematical discovery and scientific computing, SwarmEvolve finds better programs faster than text-based frameworks, and with a small open-weight LLM it improves on several results that AlphaEvolve found with frontier LLMs. These results show that large-scale search for new mathematical constructions, algorithms, and scientific computing methods does not necessarily need frontier models or large compute budgets. Swarms of small latent agents can make this kind of AI-driven scientific discovery fast, cheap, capable, and widely available.

\section*{Acknowledgments}

This work was supported by the Defense Advanced Research Projects Agency (DARPA) MATHBAC grant HR0011-26-3-E152. 

\printbibliography


\clearpage
\appendix

\section{Ablation and Scaling Studies}

\subsection{Replacing Latent Reasoning with Text Reasoning}
\label{app:text-reasoning}

Using Qwen3 8B, we replace each agent's 40 latent reasoning steps with text reasoning of up to 40 tokens, which matches the latent budget, or up to 256 tokens, which is enough for a short paragraph. The summarizer receives this text in its prompt. Latent reasoning gives the highest average score after 60 minutes, 0.804, against 0.796 and 0.767 for 40 and 256 tokens of text reasoning (Supplementary Table~\ref{tab:text-notes}). 

Text reasoning is much slower. With 256 tokens, the agents of a round take about 32 seconds to write their text reasoning, against only about 4 seconds for 40 latent reasoning steps. The summarizers then encode the text of their agents, which adds 8 seconds per round with 40 tokens and 15 seconds with 256 tokens. In 60 minutes, SwarmEvolve therefore evaluates 734 programs with latent reasoning, more than the 663 and 402 with 40 and 256 tokens of text reasoning.

Specifically, text reasoning with 40 tokens scores higher than text reasoning with 256 tokens. This is potentially due to longer reasoning messages overwhelming the $32{,}768$-token context window of the summarizer. This suggests that agent swarms can efficiently communicate with short messages, and latent reasoning and communication are more efficient than text. 

\begin{table}[H]
\centering
\caption{Best score within 60 minutes, averaged over three runs. Each agent takes 40 latent reasoning steps (Latent, the default) or writes up to 40 or 256 tokens of text reasoning, which the summarizer reads as text in its prompt. Bold: best; underlined: second best.}
\label{tab:text-notes}
\small
\renewcommand{\arraystretch}{1.15}
\begin{tabular}{ll>{\columncolor{oursShade}}ccc}
\toprule
\multicolumn{2}{l}{\footnotesize\color{black!60}Qwen3 8B $\times$ 45 agents} & \multicolumn{1}{c}{} & \multicolumn{2}{c}{\color{brown}\textbf{Text reasoning}} \\
\cmidrule(lr){4-5}
\color{brown}\textbf{Group} & \color{brown}\textbf{Problem} & \textbf{Latent} (40 steps) & 40 tokens & 256 tokens \\
\midrule
\multirow{4}{*}{Packing} & Circles in a square & \best{0.998} & \second{0.997} & 0.996 \\
  & Circles in a rectangle & \best{0.888} & 0.868 & \second{0.887} \\
  & Hexagons, $n=11$ & \best{0.627} & 0.509 & \second{0.544} \\
  & Hexagons, $n=12$ & \second{0.707} & \best{0.738} & 0.600 \\
\midrule
\multirow{5}{*}{Geometry} & Heilbronn, triangle & 0.558 & \best{0.796} & \second{0.596} \\
  & Heilbronn, convex, $n=13$ & \best{0.724} & 0.622 & \second{0.688} \\
  & Heilbronn, convex, $n=14$ & \best{0.807} & \second{0.609} & 0.536 \\
  & Distance ratio, plane & \best{0.994} & \second{0.863} & 0.828 \\
  & Distance ratio, space & \best{0.994} & \second{0.970} & 0.927 \\
\midrule
\multirow{4}{*}{Analysis} & Autocorrelation I & \second{0.995} & \best{0.996} & \second{0.995} \\
  & Autocorrelation II & \best{1.021} & \second{1.016} & \best{1.021} \\
  & Autocorrelation III & \best{1.005} & \best{1.005} & \second{1.004} \\
  & Uncertainty & \best{0.911} & \second{0.906} & \best{0.911} \\
\midrule
\multirow{3}{*}{Combinatorics} & Erd\H{o}s overlap & \best{1.000} & \second{0.999} & \second{0.999} \\
  & Sums and differences & \second{0.957} & \best{0.960} & 0.955 \\
  & Kissing number, $d=11$ & 0.309 & \second{0.320} & \best{0.400} \\
\midrule
\multirow{5}{*}{\makecell[l]{Scientific\\computing}} & Sparse linear systems & \second{0.933} & \best{1.000} & 0.775 \\
  & Eigenvalues & 0.700 & \best{0.731} & \second{0.723} \\
  & Burgers equation & \best{0.741} & \second{0.730} & 0.694 \\
  & Convection-diffusion & 0.494 & \best{0.551} & \second{0.500} \\
  & Helmholtz equation & \best{0.529} & \second{0.528} & 0.526 \\
\midrule
\multicolumn{2}{l}{Average} & \best{0.804} & \second{0.796} & 0.767 \\
\multicolumn{2}{l}{Programs evaluated per hour} & 734 & 663 & 402 \\
\bottomrule
\end{tabular}
\end{table}

\subsection{Swarm Size Scaling}
\label{app:swarm-size}

We vary the number of latent agents in each swarm with Qwen3 8B: 0, 15, 45 (the default), and 75. The novelty filter keeps two thirds of them, so each summarizer reads 0, 10, 30, or 50 latent agents. In the case of 0 latent agents, the summarizer reads the problem statement and all programs of its swarm, the union of what the latent agents would read. Each setting runs three seeds with the setup of Section~\ref{sec:experiments}.

\begin{figure}[H]
\centering
\includegraphics[width=\textwidth]{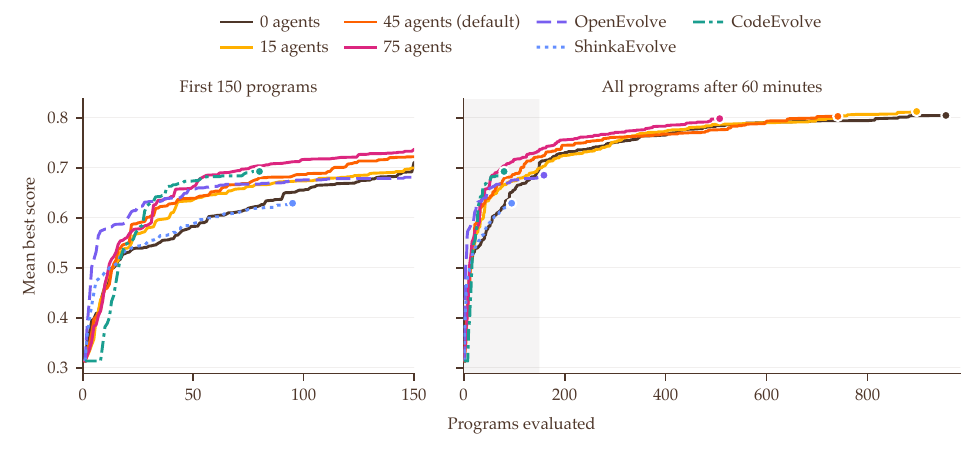}
\caption{Mean best score against the number of evaluated programs with Qwen3 8B, for 0 to 75 latent agents per swarm and the three baselines, over the first 150 programs (left) and the full 60-minute runs (right). More latent agents raise the score early in the search, and the settings converge later.}
\label{fig:agent-count-bestsofar}
\end{figure}

\begin{figure}[H]
\centering
\begin{minipage}[c]{0.5\textwidth}
\includegraphics[width=\linewidth]{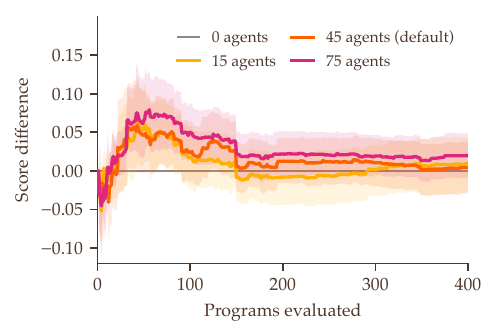}
\end{minipage}\hfill
\begin{minipage}[c]{0.46\textwidth}
\caption{Paired gain in mean best score relative to 0 latent agents per swarm, plotted against the number of evaluated programs with Qwen3 8B, with 95\% bootstrap intervals. The gains are largest in the first 100 programs and shrink within their intervals later in the search.}
\label{fig:agent-count-programs}
\end{minipage}
\end{figure}

Larger latent teams raise the score most early in the search (Supplementary Figures~\ref{fig:agent-count-bestsofar} and~\ref{fig:agent-count-programs}). Over the first 150 evaluated programs, the mean best score increases with the swarm size. From the 55th program on, SwarmEvolve with 75 agents has the highest mean best score of all settings and baselines. The three baselines evaluate 80 to 160 programs in 60 minutes, and their final scores lie below the final score of every swarm size. After about 150 programs, the gains fall within the bootstrap intervals. Smaller swarms run faster, so they evaluate more programs in the 60-minute budget and their scores keep rising for longer. All four settings end the hour at similar scores.

The early gain suggests that a larger swarm examines each program from more perspectives, so its summarizer draws on more diverse ideas. The later convergence likely reflects the limits of the LLM itself. Since all latent agents share their knowledge, additional agents can't find the blind spots that earlier agents overlook, and the diversity gain saturates. A larger model does raise the final score as shown in Table~\ref{tab:results-60min}.

\section{Time Breakdowns}
\label{app:time-breakdown}

Supplementary Table~\ref{tab:time-breakdown} divides the 60 minutes of each run among the stages of GPU work, averaged over the runs of each framework with Qwen3 8B. These times exclude the CPU time of the evaluator, which runs in parallel and never holds up the GPU. Besides proposing programs, ShinkaEvolve asks the LLM to summarize earlier programs, and CodeEvolve asks it to rewrite its system prompt. 

\begin{table}[H]
\centering
\caption{Minutes (and shares) of a 60-minute run that each framework spends on each stage of its GPU work, averaged over its runs on the 21 problems with Qwen3 8B and three seeds.}
\label{tab:time-breakdown}
\small
\renewcommand{\arraystretch}{1.2}
\begin{tabular}{l>{\columncolor{oursShade}}cccc}
\toprule
\color{brown}\textbf{Stage} & \textbf{Ours} & OpenEvolve & ShinkaEvolve & CodeEvolve \\
\midrule
Reading prompts & 7.1 min (12\%) & 1.0 min (2\%) & 0.5 min (0.9\%) & 0.1 min (0.2\%) \\
\textbf{Text reasoning} & -- & \textbf{52.5 min} (87\%) & \textbf{34.4 min} (57\%) & \textbf{46.8 min} (78\%) \\
\textbf{Latent reasoning} & \textbf{4.1 min} (7\%) & -- & -- & -- \\
Latent communication & 2.6 min (4\%) & -- & -- & -- \\
Writing programs & 39.4 min (66\%) & 6.4 min (11\%) & 5.7 min (9\%) & 4.7 min (8\%) \\
\textbf{Novelty checks} & \textbf{0.05 min} (0.1\%) & -- & \textbf{4.5 min} (7\%) & -- \\
Summarizing earlier programs & -- & -- & 14.7 min (24\%) & -- \\
Rewriting the system prompt & -- & -- & -- & 7.7 min (13\%) \\
Parameter tuning agent & 1.9 min (3\%) & -- & -- & -- \\
Other & 4.7 min (8\%) & 0.1 min (0.2\%) & 0.2 min (0.4\%) & 0.7 min (1\%) \\
\midrule
Total runtime & 60 min (100\%) & 60 min (100\%) & 60 min (100\%) & 60 min (100\%) \\
Total number of programs evaluated & 734 & 158 & 99 & 81 \\
\bottomrule
\end{tabular}
\end{table}

Compared with the three baselines, SwarmEvolve achieves a higher throughput mainly in the following two ways: 

\begin{enumerate}
  \item \textbf{Latent reasoning instead of text reasoning.} Text reasoning takes most of the GPU time of all three baselines, 20 to 35 seconds and several thousand tokens per evaluated program. SwarmEvolve replaces it with latent reasoning and latent communication, which together take about a tenth of its GPU time: the latent agents of a round reason for a few seconds in total. SwarmEvolve therefore spends most of its GPU time writing programs, and this shift explains its higher throughput.
  \item \textbf{Novelty check in latent space.} To check novelty, ShinkaEvolve embeds the code of each new program with a separate embedding model and, when its cosine similarity to an archived program exceeds 0.99, asks the LLM whether the two programs differ. These extra LLM calls take several minutes of each run. The novelty filter of SwarmEvolve instead compares the mean hidden states of the latent agents, which are already in GPU memory, and takes seconds per run.
\end{enumerate}

\section{Hyperparameters and Implementation}
\label{app:hyperparameters}

Throughout this work, our framework uses the hyperparameters in Supplementary Table~\ref{tab:hyperparameters}. The summarizers of the refinement-focused swarms, which edit one of the best programs, sample at the lowest temperature for stability, while those of the exploration-focused and fresh-start swarms, which write new approaches, sample at the highest temperature for diversity. Supplementary Table~\ref{tab:focuses} lists the focuses of the swarms.

\begin{table}[H]
\centering
\caption{Hyperparameters of SwarmEvolve used throughout all experiments in this work.}
\label{tab:hyperparameters}
\small
\renewcommand{\arraystretch}{1.15}
\begin{tabular}{llcccc}
\toprule
\multicolumn{2}{l}{\color{brown}\textbf{Agent}} & \color{brown}\textbf{Temperature} & \color{brown}\textbf{Top-$p$} & \color{brown}\textbf{Top-$k$} & \color{brown}\textbf{Maximum new tokens} \\
\midrule
\multirow{4}{*}{Summarizer} & Refinement-focused swarm & 0.3 & 0.95 & 20 & 4096 \\
  & Hybridization-focused swarm & 0.5 & 0.95 & 20 & 4096 \\
  & Exploration-focused swarm & 0.8 & 0.95 & 20 & 4096 \\
  & Fresh-start swarm & 0.8 & 0.95 & 20 & 4096 \\
\multicolumn{2}{l}{Parameter tuning agent} & 1.0 & 0.95 & 20 & 24 \\
\midrule
\multicolumn{2}{l}{\color{brown}\textbf{Hyperparameter}} & \multicolumn{4}{c}{\color{brown}\textbf{Value}} \\
\midrule
\multicolumn{2}{l}{Latent agents per swarm, run and kept} & \multicolumn{4}{c}{45, 30} \\
\multicolumn{2}{l}{Latent reasoning steps per latent agent} & \multicolumn{4}{c}{40} \\
\multicolumn{2}{l}{Maximum refinement steps of one program} & \multicolumn{4}{c}{20} \\
\multicolumn{2}{l}{One-number parameter tuning edits per round} & \multicolumn{4}{c}{up to 8} \\
\bottomrule
\end{tabular}
\end{table}

\begin{table}[H]
\centering
\caption{Focuses of the swarms. A refinement- or hybridization-focused swarm takes one of the 14 ``aspects'', and an exploration-focused or fresh-start swarm takes one of the 14 ``perspectives''. Each latent agent receives the focus in one of five phrasings, such as ``Pay particular attention to'' an aspect or ``Consider'' a perspective, followed by one of nine ways of thinking, such as ``Reason from first principles'' or ``Question one assumption the program makes'' for sample diversity.}
\label{tab:focuses}
\small
\renewcommand{\arraystretch}{1.15}
\begin{tabular}{p{0.47\textwidth}p{0.47\textwidth}}
\toprule
\color{brown}\textbf{Aspects (refinement and hybridization)} & \color{brown}\textbf{Perspectives (exploration and fresh start)} \\
\midrule
The starting point of the solution & The standard textbook method \\
The optimization or solution algorithm & A fundamentally different algorithm \\
Global exploration, restarts, and annealing & A known construction from the literature \\
Symmetries and structure of good solutions & A physics-inspired method \\
Feasibility, tolerances, and a final repair step & A parametrization that satisfies the constraints \\
The use of the time limit & Many cheap starts of one fast local optimizer \\
The largest source of the remaining error & A symmetry that forces structure \\
A smoother surrogate or a penalty formulation & A coarse-to-fine or incremental construction \\
The weakest part of the current solution & A greedy construction with local improvement \\
What the best programs in the field do differently & Linear or convex programming for convex parts \\
Bugs, wasted computation, and silent failures & An exact or analytical treatment \\
Simpler and more robust methods & Sparsity, separability, or low rank \\
The numbers in the program & A higher-order or adaptive method \\
The formulation and its free variables & An idea that nobody would try first \\
\bottomrule
\end{tabular}
\end{table}

\paragraph{Implementation.} We implement SwarmEvolve in PyTorch with the Qwen3 models of Hugging Face Transformers~\citep{wolf2020transformers}. The latent agents of a round run as one batch, and latent agents that read the same program attend to one shared copy of its KV cache through the paged FlashAttention kernel of vLLM~\citep{dao2024flashattention2,kwon2023vllm}. The summarizers of a round decode as one batch with paged KV caches and prompt-lookup speculative sampling, which drafts tokens by copying them from the programs in the context~\citep{saxena2023promptlookup} and verifies them with the acceptance rule of speculative sampling, so the samples follow the distribution of the model exactly~\citep{leviathan2023speculative}. With Qwen3 14B and 32B-FP8, SwarmEvolve also captures its decoding steps in CUDA graphs and uses the fused kernels of vLLM, including FlashAttention-3 on the H100 GPU~\citep{shah2024flashattention3}. The three baselines serve the model with vLLM, which batches requests continuously, caches shared prefixes, and runs CUDA graphs, so their decoding runs at the speed of a production serving engine. CodeEvolve runs its configuration for Qwen models with eight islands, one per CPU core. All four frameworks run Qwen3 32B-FP8, which uses the FP8 weights released by Qwen.

\section{Benchmark Problems}
\label{app:problems}

The 16 mathematical problems in Sections~\ref{app:circle-packing-square}--\ref{app:kissing} are those of Appendix~B of AlphaEvolve~\citep{novikov2025alphaevolve}, with the problem statements, initial programs, and evaluators of OpenEvolve~\citep{sharma2025openevolve}; the problem statement of circle packing in the unit square comes from ShinkaEvolve~\citep{lange2025shinkaevolve}.
All feasibility checks use the numerical tolerances of the OpenEvolve evaluators.
A construction with objective value $v$ scores $v / v_{\mathrm{AE}}$, or $v_{\mathrm{AE}} / v$ for minimization, where $v_{\mathrm{AE}}$ is the exact value of AlphaEvolve's published construction,\footnote{\url{https://github.com/google-deepmind/alphaevolve_results}, commit 4226acb.} so a score above 1 improves on AlphaEvolve.
Where an evaluator divides by a rounded $v_{\mathrm{AE}}$, we rescale its scores to the exact value.

We also design the 5 scientific computing problems in Sections~\ref{app:sparse-linear-systems}--\ref{app:helmholtz}.
Each problem consists of several cases, and a program solves one case at a time.
A case scores $\min(d_{\mathrm{cap}}, -\log_{10} e) / d_{\mathrm{cap}}$, where $e$ is the relative error with respect to a high-accuracy reference solution, and the score of a program is the mean over the cases.

\subsection{Circle Packing in the Unit Square}
\label{app:circle-packing-square}

\noindent
\begin{minipage}[t]{0.63\textwidth}
\vspace{0pt}
\paragraph{Problem.}
Circle packing asks how to place disjoint circles in a container, and this variant lets every circle have its own radius.
Pack $n = 26$ disjoint circles, with centers $c_i = (x_i, y_i)$ and radii $r_i$, inside the unit square so that the sum of their radii is as large as possible (\autoref{fig:circle-packing-square}):
\begin{align}
\max_{c_i,\, r_i} \quad & \sum_{i=1}^{n} r_i \nonumber \\
\text{subject to} \quad & r_i \le x_i \le 1 - r_i, \quad r_i \le y_i \le 1 - r_i, \nonumber \\
& r_i \ge 0, \quad \lVert c_i - c_j \rVert \ge r_i + r_j \quad \text{for } i \neq j.
\end{align}
\end{minipage}\hfill
\begin{minipage}[t]{0.33\textwidth}
\vspace{0pt}
\centering
\begin{tikzpicture}[x=3.6cm, y=3.6cm]
\draw[brown, thick] (0,0) rectangle (1,1);
\draw[plotBlue, fill=plotBlue!15] (0.8824,0.1176) circle[radius=0.1176];
\draw[plotBlue, fill=plotBlue!15] (0.1169,0.7023) circle[radius=0.1169];
\draw[plotBlue, fill=plotBlue!15] (0.6805,0.7401) circle[radius=0.0895];
\draw[plotBlue, fill=plotBlue!15] (0.1136,0.1136) circle[radius=0.1136];
\draw[plotBlue, fill=plotBlue!15] (0.3074,0.3032) circle[radius=0.1575];
\draw[plotBlue, fill=plotBlue!15] (0.3156,0.8620) circle[radius=0.1380];
\draw[plotBlue, fill=plotBlue!15] (0.5033,0.4792) circle[radius=0.1059];
\draw[plotBlue, fill=plotBlue!15] (0.9034,0.5295) circle[radius=0.0966];
\draw[plotBlue, fill=plotBlue!15] (0.5689,0.2787) circle[radius=0.1051];
\draw[plotBlue, fill=plotBlue!15] (0.3063,0.5731) circle[radius=0.1124];
\draw[plotBlue, fill=plotBlue!15] (0.0750,0.2981) circle[radius=0.0750];
\draw[plotBlue, fill=plotBlue!15] (0.6664,0.0992) circle[radius=0.0992];
\draw[plotBlue, fill=plotBlue!15] (0.9254,0.9254) circle[radius=0.0745];
\draw[plotBlue, fill=plotBlue!15] (0.4906,0.6912) circle[radius=0.1065];
\draw[plotBlue, fill=plotBlue!15] (0.4668,0.1004) circle[radius=0.1004];
\draw[plotBlue, fill=plotBlue!15] (0.8850,0.7403) circle[radius=0.1150];
\draw[plotBlue, fill=plotBlue!15] (0.0912,0.9088) circle[radius=0.0912];
\draw[plotBlue, fill=plotBlue!15] (0.1078,0.4779) circle[radius=0.1078];
\draw[plotBlue, fill=plotBlue!15] (0.7168,0.4371) circle[radius=0.1117];
\draw[plotBlue, fill=plotBlue!15] (0.7466,0.2533) circle[radius=0.0745];
\draw[plotBlue, fill=plotBlue!15] (0.5582,0.8934) circle[radius=0.1066];
\draw[plotBlue, fill=plotBlue!15] (0.9007,0.3337) circle[radius=0.0993];
\draw[plotBlue, fill=plotBlue!15] (0.7620,0.6083) circle[radius=0.0654];
\draw[plotBlue, fill=plotBlue!15] (0.6319,0.5928) circle[radius=0.0656];
\draw[plotBlue, fill=plotBlue!15] (0.7582,0.9062) circle[radius=0.0938];
\draw[plotBlue, fill=plotBlue!15] (0.2956,0.0730) circle[radius=0.0730];
\draw[plotOrange, thick] (0.3074,0.3032) -- (0.4649,0.3032);
\node[above, font=\small, text=plotOrange] at (0.3862,0.3032) {$r_i$};
\node[below, font=\small, text=brown] at (0.5,0) {$1$};
\node[left, font=\small, text=brown] at (0,0.5) {$1$};
\end{tikzpicture}
\captionsetup{type=figure}\caption{Disjoint circles in the unit square. The objective is the sum of the radii $r_i$.}
\label{fig:circle-packing-square}
\end{minipage}

\paragraph{Evaluator.}
The entry function \texttt{run\_packing()} returns the centers as a $26 \times 2$ array, the radii as an array of length 26, and their sum.
The evaluator rejects negative radii and checks both constraints with the tolerance $\epsilon = 10^{-6}$: every circle satisfies $r_i - \epsilon \le x_i \le 1 - r_i + \epsilon$ and $r_i - \epsilon \le y_i \le 1 - r_i + \epsilon$, and every pair satisfies $\lVert c_i - c_j \rVert \ge r_i + r_j - \epsilon$.
It recomputes the objective value $v = \sum_i r_i$ from the returned radii.

\subsection{Circle Packing in a Rectangle}
\label{app:circle-packing-rectangle}

\noindent
\begin{minipage}[t]{0.63\textwidth}
\vspace{0pt}
\paragraph{Problem.}
This circle packing problem fixes the perimeter of a rectangular container and lets the construction choose its shape together with the circles.
Pack $n = 21$ disjoint circles, with centers $c_i = (x_i, y_i)$ and radii $r_i$, inside a rectangle of perimeter 4 so that the sum of their radii is as large as possible.
The width $w$ of the rectangle is free, and its height is $2 - w$ (\autoref{fig:circle-packing-rectangle}):
\begin{align}
\max_{w,\, c_i,\, r_i} \quad & \sum_{i=1}^{n} r_i \nonumber \\
\text{subject to} \quad & r_i \le x_i \le w - r_i, \quad r_i \le y_i \le 2 - w - r_i, \nonumber \\
& r_i \ge 0, \quad \lVert c_i - c_j \rVert \ge r_i + r_j \quad \text{for } i \neq j.
\end{align}
\end{minipage}\hfill
\begin{minipage}[t]{0.33\textwidth}
\vspace{0pt}
\centering
\begin{tikzpicture}[x=3.223cm, y=3.223cm]
\draw[brown, thick] (0,0) rectangle (0.8829,1.1171);
\draw[plotBlue, fill=plotBlue!15] (0.5829,0.1127) circle[radius=0.1127];
\draw[plotBlue, fill=plotBlue!15] (0.5991,1.0026) circle[radius=0.1145];
\draw[plotBlue, fill=plotBlue!15] (0.3210,0.6924) circle[radius=0.1214];
\draw[plotBlue, fill=plotBlue!15] (0.2731,0.2736) circle[radius=0.0794];
\draw[plotBlue, fill=plotBlue!15] (0.1078,0.3614) circle[radius=0.1078];
\draw[plotBlue, fill=plotBlue!15] (0.1159,0.8119) circle[radius=0.1159];
\draw[plotBlue, fill=plotBlue!15] (0.5333,0.5465) circle[radius=0.1361];
\draw[plotBlue, fill=plotBlue!15] (0.7298,0.3342) circle[radius=0.1531];
\draw[plotBlue, fill=plotBlue!15] (0.6499,0.6904) circle[radius=0.0491];
\draw[plotBlue, fill=plotBlue!15] (0.1135,0.5825) circle[radius=0.1135];
\draw[plotBlue, fill=plotBlue!15] (0.0951,1.0220) circle[radius=0.0951];
\draw[plotBlue, fill=plotBlue!15] (0.3354,0.9653) circle[radius=0.1518];
\draw[plotBlue, fill=plotBlue!15] (0.4640,0.3059) circle[radius=0.1142];
\draw[plotBlue, fill=plotBlue!15] (0.3619,0.1083) circle[radius=0.1083];
\draw[plotBlue, fill=plotBlue!15] (0.7588,0.8251) circle[radius=0.1242];
\draw[plotBlue, fill=plotBlue!15] (0.7740,0.5925) circle[radius=0.1089];
\draw[plotBlue, fill=plotBlue!15] (0.7972,1.0314) circle[radius=0.0857];
\draw[plotBlue, fill=plotBlue!15] (0.3018,0.4613) circle[radius=0.1105];
\draw[plotBlue, fill=plotBlue!15] (0.1272,0.1272) circle[radius=0.1272];
\draw[plotBlue, fill=plotBlue!15] (0.7888,0.0941) circle[radius=0.0941];
\draw[plotBlue, fill=plotBlue!15] (0.5287,0.7910) circle[radius=0.1084];
\node[below, font=\small, text=brown] at (0.4415,0) {$w$};
\node[left, font=\small, text=brown] at (0,0.5585) {$2 - w$};
\end{tikzpicture}
\captionsetup{type=figure}\caption{Disjoint circles in a rectangle of width $w$ and height $2 - w$. The objective is the sum of the radii.}
\label{fig:circle-packing-rectangle}
\end{minipage}

\paragraph{Evaluator.}
The entry function \texttt{circle\_packing21()} returns a $21 \times 3$ array with rows $(x_i, y_i, r_i)$.
The evaluator rejects negative radii and computes the width $\max_i (x_i + r_i) - \min_i (x_i - r_i)$ and the height $\max_i (y_i + r_i) - \min_i (y_i - r_i)$ of the smallest axis-aligned rectangle that contains the circles.
It requires the sum of this width and height to be at most $2 + \epsilon$ and every pair of circles to satisfy $\lVert c_i - c_j \rVert \ge r_i + r_j - \epsilon$, with $\epsilon = 10^{-6}$.
The objective value is $v = \sum_i r_i$.

\subsection{Hexagon Packing with 11 Hexagons}
\label{app:hexagon-11}

\noindent
\begin{minipage}[t]{0.63\textwidth}
\vspace{0pt}
\paragraph{Problem.}
This problem packs congruent regular hexagons into the smallest regular hexagon that contains them, where each hexagon may be translated and rotated.
Let $\mathcal{H}(x, y, \theta, a)$ denote the closed regular hexagon with center $(x, y)$, side length $a$, and vertices at the angles $\theta + 60^\circ k$ for $k = 0, \dots, 5$.
Pack $n = 11$ regular hexagons of side length 1, $\mathcal{H}_i = \mathcal{H}(x_i, y_i, \theta_i, 1)$, with disjoint interiors inside an outer regular hexagon $\mathcal{H}_0 = \mathcal{H}(x_0, y_0, \theta_0, R)$ of the smallest possible side length $R$ (\autoref{fig:hexagon-11}):
\begin{align}
\min_{R,\, x_0, y_0, \theta_0,\, x_i, y_i, \theta_i} \quad & R \nonumber \\
\text{subject to} \quad & \mathcal{H}_i \subseteq \mathcal{H}_0, \quad \operatorname{int} \mathcal{H}_i \cap \operatorname{int} \mathcal{H}_j = \emptyset \quad \text{for } i \neq j,
\end{align}
where $\operatorname{int}$ denotes the interior.
\end{minipage}\hfill
\begin{minipage}[t]{0.33\textwidth}
\vspace{0pt}
\centering
\begin{tikzpicture}[x=0.47cm, y=0.47cm]
\draw[brown, thick] (3.7689,1.9987) -- (0.7641,4.9004) -- (-3.2513,3.7490) -- (-4.2619,-0.3041) -- (-1.2571,-3.2058) -- (2.7583,-2.0544) -- cycle;
\draw[plotPurple, fill=plotPurple!15] (1.0000,0.0000) -- (0.5000,0.8660) -- (-0.5000,0.8660) -- (-1.0000,0.0000) -- (-0.5000,-0.8660) -- (0.5000,-0.8660) -- cycle;
\draw[plotPurple, fill=plotPurple!15] (1.0000,1.8360) -- (0.5000,2.7020) -- (-0.5000,2.7020) -- (-1.0000,1.8360) -- (-0.5000,0.9699) -- (0.5000,0.9699) -- cycle;
\draw[plotPurple, fill=plotPurple!15] (2.5900,0.9180) -- (2.0900,1.7840) -- (1.0900,1.7840) -- (0.5900,0.9180) -- (1.0900,0.0520) -- (2.0900,0.0520) -- cycle;
\draw[plotPurple, fill=plotPurple!15] (-0.5900,0.9180) -- (-1.0900,1.7840) -- (-2.0900,1.7840) -- (-2.5900,0.9180) -- (-2.0900,0.0520) -- (-1.0900,0.0520) -- cycle;
\draw[plotPurple, fill=plotPurple!15] (2.5900,-0.9180) -- (2.0900,-0.0520) -- (1.0900,-0.0520) -- (0.5900,-0.9180) -- (1.0900,-1.7840) -- (2.0900,-1.7840) -- cycle;
\draw[plotPurple, fill=plotPurple!15] (-0.5900,-0.9180) -- (-1.0900,-0.0520) -- (-2.0900,-0.0520) -- (-2.5900,-0.9180) -- (-2.0900,-1.7840) -- (-1.0900,-1.7840) -- cycle;
\draw[plotPurple, fill=plotPurple!15] (1.0000,-1.8360) -- (0.5000,-0.9699) -- (-0.5000,-0.9699) -- (-1.0000,-1.8360) -- (-0.5000,-2.7020) -- (0.5000,-2.7020) -- cycle;
\draw[plotPurple, fill=plotPurple!15] (2.5900,2.7540) -- (2.0900,3.6200) -- (1.0900,3.6200) -- (0.5900,2.7540) -- (1.0900,1.8879) -- (2.0900,1.8879) -- cycle;
\draw[plotPurple, fill=plotPurple!15] (-0.5900,2.7540) -- (-1.0900,3.6200) -- (-2.0900,3.6200) -- (-2.5900,2.7540) -- (-2.0900,1.8879) -- (-1.0900,1.8879) -- cycle;
\draw[plotPurple, fill=plotPurple!15] (1.0000,3.6719) -- (0.5000,4.5380) -- (-0.5000,4.5380) -- (-1.0000,3.6719) -- (-0.5000,2.8059) -- (0.5000,2.8059) -- cycle;
\draw[plotPurple, fill=plotPurple!15] (-2.1800,0.0000) -- (-2.6800,0.8660) -- (-3.6800,0.8660) -- (-4.1800,0.0000) -- (-3.6800,-0.8660) -- (-2.6800,-0.8660) -- cycle;
\node[above, font=\small, text=brown] at (-1.2436,4.3247) {$R$};
\end{tikzpicture}
\captionsetup{type=figure}\caption{Eleven disjoint unit hexagons inside a regular hexagon of side length $R$, which is minimized.}
\label{fig:hexagon-11}
\end{minipage}

\paragraph{Evaluator.}
The entry function \texttt{hexagon\_packing\_11()} returns an $11 \times 3$ array with rows $(x_i, y_i, \theta_i)$, with angles in degrees, the triple $(x_0, y_0, \theta_0)$ of the outer hexagon, and $R$.
The evaluator computes the six vertices of every hexagon.
For every pair of inner hexagons, it projects both hexagons onto the edge normals of the two hexagons and requires a gap larger than $\epsilon$ along at least one normal, which is the separating axis test for convex polygons.
It requires every vertex of every inner hexagon to lie inside the outer hexagon, with the tolerance $\epsilon$ on its distance to each edge, where $\epsilon = 10^{-6}$.
The objective value is $v = R$.

\subsection{Hexagon Packing with 12 Hexagons}
\label{app:hexagon-12}

\noindent
\begin{minipage}[t]{0.63\textwidth}
\vspace{0pt}
\paragraph{Problem.}
This problem packs congruent regular hexagons into the smallest regular hexagon that contains them, where each hexagon may be translated and rotated.
Let $\mathcal{H}(x, y, \theta, a)$ denote the closed regular hexagon with center $(x, y)$, side length $a$, and vertices at the angles $\theta + 60^\circ k$ for $k = 0, \dots, 5$.
Pack $n = 12$ regular hexagons of side length 1, $\mathcal{H}_i = \mathcal{H}(x_i, y_i, \theta_i, 1)$, with disjoint interiors inside an outer regular hexagon $\mathcal{H}_0 = \mathcal{H}(x_0, y_0, \theta_0, R)$ of the smallest possible side length $R$ (\autoref{fig:hexagon-12}):
\begin{align}
\min_{R,\, x_0, y_0, \theta_0,\, x_i, y_i, \theta_i} \quad & R \nonumber \\
\text{subject to} \quad & \mathcal{H}_i \subseteq \mathcal{H}_0, \quad \operatorname{int} \mathcal{H}_i \cap \operatorname{int} \mathcal{H}_j = \emptyset \quad \text{for } i \neq j,
\end{align}
where $\operatorname{int}$ denotes the interior.
\end{minipage}\hfill
\begin{minipage}[t]{0.33\textwidth}
\vspace{0pt}
\centering
\begin{tikzpicture}[x=0.47cm, y=0.47cm]
\draw[brown, thick] (4.1800,0.0000) -- (2.0900,3.6200) -- (-2.0900,3.6200) -- (-4.1800,0.0000) -- (-2.0900,-3.6200) -- (2.0900,-3.6200) -- cycle;
\draw[plotPurple, fill=plotPurple!15] (1.0000,0.0000) -- (0.5000,0.8660) -- (-0.5000,0.8660) -- (-1.0000,0.0000) -- (-0.5000,-0.8660) -- (0.5000,-0.8660) -- cycle;
\draw[plotPurple, fill=plotPurple!15] (2.5900,0.9180) -- (2.0900,1.7840) -- (1.0900,1.7840) -- (0.5900,0.9180) -- (1.0900,0.0520) -- (2.0900,0.0520) -- cycle;
\draw[plotPurple, fill=plotPurple!15] (2.5900,-0.9180) -- (2.0900,-0.0520) -- (1.0900,-0.0520) -- (0.5900,-0.9180) -- (1.0900,-1.7840) -- (2.0900,-1.7840) -- cycle;
\draw[plotPurple, fill=plotPurple!15] (1.0000,1.8360) -- (0.5000,2.7020) -- (-0.5000,2.7020) -- (-1.0000,1.8360) -- (-0.5000,0.9699) -- (0.5000,0.9699) -- cycle;
\draw[plotPurple, fill=plotPurple!15] (-0.5900,0.9180) -- (-1.0900,1.7840) -- (-2.0900,1.7840) -- (-2.5900,0.9180) -- (-2.0900,0.0520) -- (-1.0900,0.0520) -- cycle;
\draw[plotPurple, fill=plotPurple!15] (-0.5900,-0.9180) -- (-1.0900,-0.0520) -- (-2.0900,-0.0520) -- (-2.5900,-0.9180) -- (-2.0900,-1.7840) -- (-1.0900,-1.7840) -- cycle;
\draw[plotPurple, fill=plotPurple!15] (1.0000,-1.8360) -- (0.5000,-0.9699) -- (-0.5000,-0.9699) -- (-1.0000,-1.8360) -- (-0.5000,-2.7020) -- (0.5000,-2.7020) -- cycle;
\draw[plotPurple, fill=plotPurple!15] (4.1800,0.0000) -- (3.6800,0.8660) -- (2.6800,0.8660) -- (2.1800,0.0000) -- (2.6800,-0.8660) -- (3.6800,-0.8660) -- cycle;
\draw[plotPurple, fill=plotPurple!15] (-2.1800,0.0000) -- (-2.6800,0.8660) -- (-3.6800,0.8660) -- (-4.1800,0.0000) -- (-3.6800,-0.8660) -- (-2.6800,-0.8660) -- cycle;
\draw[plotPurple, fill=plotPurple!15] (2.5900,-2.7540) -- (2.0900,-1.8879) -- (1.0900,-1.8879) -- (0.5900,-2.7540) -- (1.0900,-3.6200) -- (2.0900,-3.6200) -- cycle;
\draw[plotPurple, fill=plotPurple!15] (2.5900,2.7540) -- (2.0900,3.6200) -- (1.0900,3.6200) -- (0.5900,2.7540) -- (1.0900,1.8879) -- (2.0900,1.8879) -- cycle;
\draw[plotPurple, fill=plotPurple!15] (-0.5900,2.7540) -- (-1.0900,3.6200) -- (-2.0900,3.6200) -- (-2.5900,2.7540) -- (-2.0900,1.8879) -- (-1.0900,1.8879) -- cycle;
\node[above, font=\small, text=brown] at (0.0000,3.6200) {$R$};
\end{tikzpicture}
\captionsetup{type=figure}\caption{Twelve disjoint unit hexagons inside a regular hexagon of side length $R$, which is minimized.}
\label{fig:hexagon-12}
\end{minipage}

\paragraph{Evaluator.}
The entry function \texttt{hexagon\_packing\_12()} returns a $12 \times 3$ array with rows $(x_i, y_i, \theta_i)$, with angles in degrees, the triple $(x_0, y_0, \theta_0)$ of the outer hexagon, and $R$.
The evaluator computes the six vertices of every hexagon.
For every pair of inner hexagons, it projects both hexagons onto the edge normals of the two hexagons and requires a gap larger than $\epsilon$ along at least one normal, which is the separating axis test for convex polygons.
It requires every vertex of every inner hexagon to lie inside the outer hexagon, with the tolerance $\epsilon$ on its distance to each edge, where $\epsilon = 10^{-6}$.
The objective value is $v = R$.

\subsection{Heilbronn Problem in a Triangle}
\label{app:heilbronn-triangle}

\noindent
\begin{minipage}[t]{0.63\textwidth}
\vspace{0pt}
\paragraph{Problem.}
The Heilbronn problem asks for $n$ points in a region that maximize the smallest area of a triangle formed by three of the points~\citep{roth1951heilbronn,comellas2002heilbronn}.
Place $n = 11$ points in the equilateral triangle $T$ with vertices $(0, 0)$, $(1, 0)$, and $(\tfrac12, \tfrac{\sqrt3}{2})$ to maximize the smallest area of a triangle formed by three of the points (\autoref{fig:heilbronn-triangle}):
\begin{align}
\max_{p_1, \dots, p_n \in T} \; \frac{\min_{i < j < k} \operatorname{area}(p_i, p_j, p_k)}{\operatorname{area}(T)},
\end{align}
where $\operatorname{area}(p_i, p_j, p_k)$ is the area of the triangle with vertices $p_i$, $p_j$, and $p_k$, and $\operatorname{area}(T) = \sqrt3/4$.
\end{minipage}\hfill
\begin{minipage}[t]{0.33\textwidth}
\vspace{0pt}
\centering
\begin{tikzpicture}[x=3.8cm, y=3.8cm]
\draw[brown, thick] (0,0) -- (1,0) -- (0.5,0.8660) -- cycle;
\draw[plotMagenta, fill=plotMagenta!30] (0.7519,0.2427) -- (0.3973,0.5834) -- (0.5503,0.3606) -- cycle;
\fill[plotBlue] (0.1407,0.1389) circle[radius=1.8pt];
\fill[plotBlue] (0.1433,0.2482) circle[radius=1.8pt];
\fill[plotBlue] (0.9256,0.0000) circle[radius=1.8pt];
\fill[plotBlue] (0.5829,0.7224) circle[radius=1.8pt];
\fill[plotBlue] (0.7519,0.2427) circle[radius=1.8pt];
\fill[plotBlue] (0.4434,-0.0000) circle[radius=1.8pt];
\fill[plotBlue] (0.3973,0.5834) circle[radius=1.8pt];
\fill[plotBlue] (0.3865,0.1367) circle[radius=1.8pt];
\fill[plotBlue] (0.4628,0.8016) circle[radius=1.8pt];
\fill[plotBlue] (0.7915,0.3611) circle[radius=1.8pt];
\fill[plotBlue] (0.5503,0.3606) circle[radius=1.8pt];
\end{tikzpicture}
\captionsetup{type=figure}\caption{Eleven points in the equilateral triangle. The shaded triangle has the smallest area among all triangles formed by the points.}
\label{fig:heilbronn-triangle}
\end{minipage}

\paragraph{Evaluator.}
The entry function \texttt{heilbronn\_triangle11()} returns an $11 \times 2$ array of points.
The evaluator checks that every point $(x, y)$ satisfies the three edge inequalities of $T$, $y \ge -\epsilon$, $\sqrt3 x \le \sqrt3 - y + \epsilon$, and $y \le \sqrt3 x + \epsilon$, with $\epsilon = 10^{-6}$.
It computes the areas $\tfrac12 \lvert \det(p_j - p_i, p_k - p_i) \rvert$ of all 165 triangles and divides the smallest area by $\sqrt3/4$ to obtain $v$.

\subsection{Heilbronn Problem in a Convex Region with 13 Points}
\label{app:heilbronn-convex-13}

\noindent
\begin{minipage}[t]{0.63\textwidth}
\vspace{0pt}
\paragraph{Problem.}
The Heilbronn problem asks for $n$ points in a region that maximize the smallest area of a triangle formed by three of the points~\citep{roth1951heilbronn,comellas2002heilbronn}.
In this variant the region is any convex region of unit area, so the relevant region is the convex hull of the points, and the ratio below is invariant under invertible affine maps of the plane.
Place $n = 13$ points in the plane to maximize the smallest area of a triangle formed by three of the points, relative to the area of their convex hull (\autoref{fig:heilbronn-convex-13}):
\begin{align}
\max_{p_1, \dots, p_n \in \mathbb{R}^2} \; \frac{\min_{i < j < k} \operatorname{area}(p_i, p_j, p_k)}{\operatorname{area}(\operatorname{conv}\{p_1, \dots, p_n\})},
\end{align}
where $\operatorname{area}(p_i, p_j, p_k)$ is the area of the triangle with vertices $p_i$, $p_j$, and $p_k$, and $\operatorname{conv}$ denotes the convex hull.
\end{minipage}\hfill
\begin{minipage}[t]{0.33\textwidth}
\vspace{0pt}
\centering
\begin{tikzpicture}[x=1.85cm, y=1.85cm]
\draw[brown, thick, dashed, fill=plotYellow!12] (0.9345,-0.5757) -- (0.9565,-0.2917) -- (0.6892,0.2663) -- (0.3228,0.7009) -- (-0.1323,1.0000) -- (-0.5279,0.4678) -- (-0.8137,-0.1398) -- (-0.8884,-0.6072) -- (0.0255,-0.6402) -- cycle;
\draw[plotMagenta, fill=plotMagenta!30] (0.0033,0.7170) -- (0.3228,0.7009) -- (-0.1323,1.0000) -- cycle;
\fill[plotBlue] (-0.3905,0.0333) circle[radius=1.8pt];
\fill[plotBlue] (-0.5279,0.4678) circle[radius=1.8pt];
\fill[plotBlue] (-0.8137,-0.1398) circle[radius=1.8pt];
\fill[plotBlue] (-0.4340,-0.4713) circle[radius=1.8pt];
\fill[plotBlue] (0.0033,0.7170) circle[radius=1.8pt];
\fill[plotBlue] (0.3228,0.7009) circle[radius=1.8pt];
\fill[plotBlue] (0.0255,-0.6402) circle[radius=1.8pt];
\fill[plotBlue] (-0.1323,1.0000) circle[radius=1.8pt];
\fill[plotBlue] (0.9565,-0.2917) circle[radius=1.8pt];
\fill[plotBlue] (0.2551,-0.4594) circle[radius=1.8pt];
\fill[plotBlue] (-0.8884,-0.6072) circle[radius=1.8pt];
\fill[plotBlue] (0.6892,0.2663) circle[radius=1.8pt];
\fill[plotBlue] (0.9345,-0.5757) circle[radius=1.8pt];
\end{tikzpicture}
\captionsetup{type=figure}\caption{Thirteen points with their convex hull (dashed). The shaded triangle has the smallest area.}
\label{fig:heilbronn-convex-13}
\end{minipage}

\paragraph{Evaluator.}
The entry function \texttt{heilbronn\_convex13()} returns a $13 \times 2$ array of points anywhere in the plane.
The evaluator computes the areas of all 286 triangles and the area of the convex hull of the points with the Quickhull algorithm~\citep{barber1996quickhull} in SciPy.
It divides the smallest triangle area by the hull area to obtain $v$.

\subsection{Heilbronn Problem in a Convex Region with 14 Points}
\label{app:heilbronn-convex-14}

\noindent
\begin{minipage}[t]{0.63\textwidth}
\vspace{0pt}
\paragraph{Problem.}
The Heilbronn problem asks for $n$ points in a region that maximize the smallest area of a triangle formed by three of the points~\citep{roth1951heilbronn,comellas2002heilbronn}.
In this variant the region is any convex region of unit area, so the relevant region is the convex hull of the points, and the ratio below is invariant under invertible affine maps of the plane.
Place $n = 14$ points in the plane to maximize the smallest area of a triangle formed by three of the points, relative to the area of their convex hull (\autoref{fig:heilbronn-convex-14}):
\begin{align}
\max_{p_1, \dots, p_n \in \mathbb{R}^2} \; \frac{\min_{i < j < k} \operatorname{area}(p_i, p_j, p_k)}{\operatorname{area}(\operatorname{conv}\{p_1, \dots, p_n\})},
\end{align}
where $\operatorname{area}(p_i, p_j, p_k)$ is the area of the triangle with vertices $p_i$, $p_j$, and $p_k$, and $\operatorname{conv}$ denotes the convex hull.
\end{minipage}\hfill
\begin{minipage}[t]{0.33\textwidth}
\vspace{0pt}
\centering
\begin{tikzpicture}[x=1.85cm, y=1.85cm]
\draw[brown, thick, dashed, fill=plotYellow!12] (-0.6288,0.8653) -- (-1.0000,-0.5690) -- (-0.2576,-0.9044) -- (0.2078,-0.9835) -- (0.5339,-0.8295) -- (0.7242,-0.4408) -- (0.8494,0.3271) -- (0.7175,0.7871) -- (0.0948,0.9230) -- cycle;
\draw[plotMagenta, fill=plotMagenta!30] (0.7242,-0.4408) -- (0.0948,0.9230) -- (0.4959,0.2086) -- cycle;
\fill[plotBlue] (-1.0000,-0.5690) circle[radius=1.8pt];
\fill[plotBlue] (0.7242,-0.4408) circle[radius=1.8pt];
\fill[plotBlue] (0.2078,-0.9835) circle[radius=1.8pt];
\fill[plotBlue] (0.5339,-0.8295) circle[radius=1.8pt];
\fill[plotBlue] (-0.2576,-0.9044) circle[radius=1.8pt];
\fill[plotBlue] (0.0948,0.9230) circle[radius=1.8pt];
\fill[plotBlue] (-0.4163,-0.4103) circle[radius=1.8pt];
\fill[plotBlue] (-0.6288,0.8653) circle[radius=1.8pt];
\fill[plotBlue] (0.8494,0.3271) circle[radius=1.8pt];
\fill[plotBlue] (-0.7672,0.0680) circle[radius=1.8pt];
\fill[plotBlue] (-0.3767,0.1915) circle[radius=1.8pt];
\fill[plotBlue] (0.4959,0.2086) circle[radius=1.8pt];
\fill[plotBlue] (-0.1768,0.7667) circle[radius=1.8pt];
\fill[plotBlue] (0.7175,0.7871) circle[radius=1.8pt];
\end{tikzpicture}
\captionsetup{type=figure}\caption{Fourteen points with their convex hull (dashed). The shaded triangle has the smallest area.}
\label{fig:heilbronn-convex-14}
\end{minipage}

\paragraph{Evaluator.}
The entry function \texttt{heilbronn\_convex14()} returns a $14 \times 2$ array of points anywhere in the plane.
The evaluator computes the areas of all 364 triangles and the area of the convex hull of the points with the Quickhull algorithm~\citep{barber1996quickhull} in SciPy.
It divides the smallest triangle area by the hull area to obtain $v$.

\subsection{Distance Ratio in the Plane}
\label{app:distance-ratio-plane}

\noindent
\begin{minipage}[t]{0.63\textwidth}
\vspace{0pt}
\paragraph{Problem.}
This problem asks for evenly spread point sets, whose smallest pairwise distance is large relative to the diameter of the set~\citep{bateman1951geometrical}.
Place $n = 16$ points in the plane to maximize the ratio of their smallest pairwise distance $d_{\min} = \min_{i \neq j} \lVert p_i - p_j \rVert$ to their largest pairwise distance $d_{\max} = \max_{i \neq j} \lVert p_i - p_j \rVert$ (\autoref{fig:distance-ratio-plane}):
\begin{align}
\max_{p_1, \dots, p_n \in \mathbb{R}^2} \; \left( \frac{d_{\min}}{d_{\max}} \right)^2.
\end{align}
The ratio is invariant under translations, rotations, and scalings of the point set.
\end{minipage}\hfill
\begin{minipage}[t]{0.33\textwidth}
\vspace{0pt}
\centering
\begin{tikzpicture}[x=3.6cm, y=3.6cm]
\draw[brown!40] (0.3431,0.3672) -- (0.0912,0.4860);
\draw[brown!40] (0.3431,0.3672) -- (0.4891,0.1300);
\draw[brown!40] (-0.1870,0.4722) -- (-0.4034,0.2968);
\draw[brown!40] (-0.1870,0.4722) -- (0.0912,0.4860);
\draw[brown!40] (0.0086,0.2200) -- (0.0912,0.4860);
\draw[brown!40] (0.0086,0.2200) -- (0.2242,0.0437);
\draw[brown!40] (0.0086,0.2200) -- (-0.2310,0.0780);
\draw[brown!40] (-0.4034,0.2968) -- (-0.5067,0.0382);
\draw[brown!40] (-0.4034,0.2968) -- (-0.2310,0.0780);
\draw[brown!40] (-0.4428,-0.2329) -- (-0.5067,0.0382);
\draw[brown!40] (-0.4428,-0.2329) -- (-0.1671,-0.1931);
\draw[brown!40] (0.1109,-0.2107) -- (0.3285,-0.3846);
\draw[brown!40] (0.1109,-0.2107) -- (0.2242,0.0437);
\draw[brown!40] (0.1109,-0.2107) -- (-0.1671,-0.1931);
\draw[brown!40] (0.0721,-0.4936) -- (0.3285,-0.3846);
\draw[brown!40] (0.0721,-0.4936) -- (-0.2053,-0.4690);
\draw[brown!40] (0.4758,-0.1482) -- (0.3285,-0.3846);
\draw[brown!40] (0.4758,-0.1482) -- (0.4891,0.1300);
\draw[brown!40] (-0.5067,0.0382) -- (-0.2310,0.0780);
\draw[brown!40] (0.2242,0.0437) -- (0.4891,0.1300);
\draw[brown!40] (-0.2310,0.0780) -- (-0.1671,-0.1931);
\draw[brown!40] (-0.2053,-0.4690) -- (-0.1671,-0.1931);
\draw[plotPurple, thick, dashed] (0.3431,0.3672) -- node[pos=0.35, fill=white, inner sep=1pt, font=\small, text=plotPurple] {$d_{\max}$} (-0.2053,-0.4690);
\draw[plotOrange, very thick] (-0.2310,0.0780) -- (-0.1671,-0.1931);
\fill[plotBlue] (0.3431,0.3672) circle[radius=1.8pt];
\fill[plotBlue] (-0.1870,0.4722) circle[radius=1.8pt];
\fill[plotBlue] (0.0086,0.2200) circle[radius=1.8pt];
\fill[plotBlue] (-0.4034,0.2968) circle[radius=1.8pt];
\fill[plotBlue] (-0.4428,-0.2329) circle[radius=1.8pt];
\fill[plotBlue] (0.1109,-0.2107) circle[radius=1.8pt];
\fill[plotBlue] (0.0721,-0.4936) circle[radius=1.8pt];
\fill[plotBlue] (0.4758,-0.1482) circle[radius=1.8pt];
\fill[plotBlue] (-0.5067,0.0382) circle[radius=1.8pt];
\fill[plotBlue] (0.3285,-0.3846) circle[radius=1.8pt];
\fill[plotBlue] (0.0912,0.4860) circle[radius=1.8pt];
\fill[plotBlue] (0.2242,0.0437) circle[radius=1.8pt];
\fill[plotBlue] (-0.2310,0.0780) circle[radius=1.8pt];
\fill[plotBlue] (-0.2053,-0.4690) circle[radius=1.8pt];
\fill[plotBlue] (-0.1671,-0.1931) circle[radius=1.8pt];
\fill[plotBlue] (0.4891,0.1300) circle[radius=1.8pt];
\node[font=\small, text=plotOrange] at (-0.2867,-0.0782) {$d_{\min}$};
\end{tikzpicture}
\captionsetup{type=figure}\caption{Sixteen points in the plane. The thin lines join the pairs at the smallest distance $d_{\min}$, and the dashed line marks the largest distance $d_{\max}$.}
\label{fig:distance-ratio-plane}
\end{minipage}

\paragraph{Evaluator.}
The entry function \texttt{min\_max\_dist\_dim2\_16()} returns a $16 \times 2$ array of points anywhere in the plane.
The evaluator computes the 120 pairwise distances with SciPy and forms $v = (d_{\min}/d_{\max})^2$.

\subsection{Distance Ratio in Space}
\label{app:distance-ratio-space}

\noindent
\begin{minipage}[t]{0.63\textwidth}
\vspace{0pt}
\paragraph{Problem.}
This problem asks for evenly spread point sets, whose smallest pairwise distance is large relative to the diameter of the set~\citep{bateman1951geometrical}.
Place $n = 14$ points in three-dimensional space to maximize the ratio of their smallest pairwise distance $d_{\min} = \min_{i \neq j} \lVert p_i - p_j \rVert$ to their largest pairwise distance $d_{\max} = \max_{i \neq j} \lVert p_i - p_j \rVert$ (\autoref{fig:distance-ratio-space}):
\begin{align}
\max_{p_1, \dots, p_n \in \mathbb{R}^3} \; \left( \frac{d_{\min}}{d_{\max}} \right)^2.
\end{align}
The ratio is invariant under translations, rotations, and scalings of the point set.
\end{minipage}\hfill
\begin{minipage}[t]{0.33\textwidth}
\vspace{0pt}
\centering
\begin{tikzpicture}[x=3.6cm, y=3.6cm]
\draw[brown!27] (0.0413,-0.3533) -- (0.2257,0.0906);
\draw[brown!30] (-0.3261,-0.0293) -- (0.0413,-0.3533);
\draw[brown!31] (0.2257,0.0906) -- (0.1235,0.5020);
\draw[brown!34] (0.4704,-0.1035) -- (0.2257,0.0906);
\draw[brown!36] (-0.3261,-0.0293) -- (-0.3398,0.3894);
\draw[brown!38] (0.0413,-0.3533) -- (-0.3011,-0.3814);
\draw[brown!39] (-0.3261,-0.0293) -- (-0.3011,-0.3814);
\draw[brown!39] (0.0009,0.0037) -- (-0.3261,-0.0293);
\draw[brown!40] (0.1235,0.5020) -- (-0.3398,0.3894);
\draw[brown!41] (0.0413,-0.3533) -- (0.1710,-0.4469);
\draw[brown!46] (0.0009,0.0037) -- (0.4704,-0.1035);
\draw[brown!48] (0.0009,0.0037) -- (-0.3011,-0.3814);
\draw[brown!48] (0.3223,0.3664) -- (0.1235,0.5020);
\draw[brown!48] (0.4704,-0.1035) -- (0.1710,-0.4469);
\draw[brown!50] (-0.3011,-0.3814) -- (0.1710,-0.4469);
\draw[brown!50] (-0.4830,0.0236) -- (-0.3398,0.3894);
\draw[brown!51] (0.0009,0.0037) -- (0.1710,-0.4469);
\draw[brown!53] (-0.3011,-0.3814) -- (-0.4830,0.0236);
\draw[brown!56] (0.4704,-0.1035) -- (0.3383,-0.0903);
\draw[brown!58] (-0.3011,-0.3814) -- (-0.1303,-0.2303);
\draw[brown!59] (0.0009,0.0037) -- (-0.1132,0.2593);
\draw[brown!59] (0.0009,0.0037) -- (-0.1303,-0.2303);
\draw[brown!61] (0.1710,-0.4469) -- (0.3383,-0.0903);
\draw[brown!61] (-0.1303,-0.2303) -- (0.1710,-0.4469);
\draw[brown!63] (-0.1132,0.2593) -- (-0.4830,0.0236);
\draw[brown!64] (0.3223,0.3664) -- (0.3383,-0.0903);
\draw[brown!64] (-0.1303,-0.2303) -- (-0.4830,0.0236);
\draw[brown!64] (0.3223,0.3664) -- (-0.1132,0.2593);
\draw[brown!69] (-0.1303,-0.2303) -- (0.3383,-0.0903);
\draw[brown!69] (-0.1132,0.2593) -- (-0.1303,-0.2303);
\draw[plotPurple, thick, dashed] (0.2257,0.0906) -- node[pos=0.3, fill=white, inner sep=1pt, font=\small, text=plotPurple] {$d_{\max}$} (-0.1303,-0.2303);
\fill[plotBlue!35] (0.2257,0.0906) circle[radius=1.30pt];
\fill[plotBlue!42] (0.0413,-0.3533) circle[radius=1.40pt];
\fill[plotBlue!42] (-0.3261,-0.0293) circle[radius=1.41pt];
\fill[plotBlue!53] (0.1235,0.5020) circle[radius=1.55pt];
\fill[plotBlue!61] (-0.3398,0.3894) circle[radius=1.67pt];
\fill[plotBlue!62] (0.4704,-0.1035) circle[radius=1.69pt];
\fill[plotBlue!67] (-0.3011,-0.3814) circle[radius=1.76pt];
\fill[plotBlue!69] (0.0009,0.0037) circle[radius=1.78pt];
\fill[plotBlue!76] (0.1710,-0.4469) circle[radius=1.87pt];
\fill[plotBlue!83] (-0.4830,0.0236) circle[radius=1.97pt];
\fill[plotBlue!84] (0.3223,0.3664) circle[radius=1.99pt];
\fill[plotBlue!97] (0.3383,-0.0903) circle[radius=2.17pt];
\fill[plotBlue!99] (-0.1132,0.2593) circle[radius=2.19pt];
\fill[plotBlue!100] (-0.1303,-0.2303) circle[radius=2.20pt];
\end{tikzpicture}
\captionsetup{type=figure}\caption{Fourteen points in space, shaded by depth. The thin lines join the pairs at the smallest distance $d_{\min}$, and the dashed line marks the largest distance $d_{\max}$.}
\label{fig:distance-ratio-space}
\end{minipage}

\paragraph{Evaluator.}
The entry function \texttt{min\_max\_dist\_dim3\_14()} returns a $14 \times 3$ array of points anywhere in space.
The evaluator computes the 91 pairwise distances with SciPy and forms $v = (d_{\min}/d_{\max})^2$.

\subsection{First Autocorrelation Inequality}
\label{app:first-autocorrelation}

\paragraph{Problem.}
For every non-negative function $f$ supported on $[-\tfrac14, \tfrac14]$, the first autocorrelation inequality states
\begin{align}
\max_{-1/2 \le t \le 1/2} (f \ast f)(t) \ge C_1 \left( \int_{-1/4}^{1/4} f(x) \, dx \right)^2,
\end{align}
where $(f \ast f)(t) = \int f(t - x) f(x) \, dx$ is the autoconvolution of $f$.
The constant $C_1$ bounds the size of Sidon sets in additive combinatorics, and the bounds $1.28 \le C_1 \le 1.5098$ come from \citet{cloninger2017suprema} and \citet{matolcsi2010improved}.
Every such $f$ with a positive integral gives an upper bound on $C_1$.
Find a non-negative function $f$ supported on $[-\tfrac14, \tfrac14]$ with a positive integral that minimizes this bound,
\begin{align}
\min_{f \ge 0} \; \frac{\max_t (f \ast f)(t)}{\left( \int f \right)^2}.
\end{align}
The program represents $f$ by its values $f_1, \dots, f_N$ on $N$ equal intervals of $[-\tfrac14, \tfrac14]$ and chooses $N$.

\paragraph{Evaluator.}
The entry function \texttt{run()} returns $f_1, \dots, f_N$, the bound $\tilde v$, a loss value, and $N$.
The evaluator requires every $f_i$ to be non-negative and, with the interval width $\Delta = 1/(2N)$, computes
\begin{align}
v = \frac{\max_k \Delta \sum_i f_i f_{k-i}}{\left( \Delta \sum_i f_i \right)^2}.
\end{align}
The autoconvolution of a step function is piecewise linear with breakpoints at multiples of $\Delta$, so $v$ is the exact bound of the step function with values $f_i$.
The evaluator requires $(\Delta \sum_i f_i)^2 \ge 10^{-8}$ and $\lvert v - \tilde v \rvert \le 10^{-6}$, and the score uses $\tilde v$.

\subsection{Second Autocorrelation Inequality}
\label{app:second-autocorrelation}

\paragraph{Problem.}
For every non-negative function $f$, the second autocorrelation inequality states
\begin{align}
\lVert f \ast f \rVert_2^2 \le C_2 \lVert f \ast f \rVert_1 \lVert f \ast f \rVert_\infty,
\end{align}
where $(f \ast f)(t) = \int f(t - x) f(x) \, dx$ is the autoconvolution of $f$ and $\lVert f \ast f \rVert_1 = (\int f)^2$~\citep{martin2009supremum}.
The inequality $\lVert g \rVert_2^2 \le \lVert g \rVert_1 \lVert g \rVert_\infty$ gives $C_2 \le 1$, and \citet{matolcsi2010improved} prove $C_2 \ge 0.88922$ with an explicit function.
Every such $f$ with a positive integral gives a lower bound on $C_2$.
Find a non-negative function $f$ with a positive integral that maximizes this bound,
\begin{align}
\max_{f \ge 0} \; \frac{\lVert f \ast f \rVert_2^2}{\lVert f \ast f \rVert_1 \lVert f \ast f \rVert_\infty}.
\end{align}
The program represents $f$ by its values $f_1, \dots, f_N$ on $N$ equal intervals and chooses $N$.

\paragraph{Evaluator.}
The entry function \texttt{run()} returns $f_1, \dots, f_N$, the bound $\tilde v$, a loss value, and $N$.
The evaluator requires $f_i \ge -10^{-6}$, sets negative values to zero, and forms the discrete autoconvolution $g_k = \sum_i f_i f_{k-i}$ for $k = 1, \dots, 2N - 1$.
It places these values on the interior nodes of a uniform grid with spacing $\Delta = 1/(2N)$ on $[-\tfrac12, \tfrac12]$, sets $g_0 = g_{2N} = 0$ at the two end nodes, and computes the norms of the piecewise linear interpolant $g$ of these values exactly,
\begin{align}
\lVert g \rVert_2^2 &= \sum_{k=0}^{2N-1} \frac{\Delta}{3} \left( g_k^2 + g_k g_{k+1} + g_{k+1}^2 \right), &
\lVert g \rVert_1 &= \Delta \sum_{k} \lvert g_k \rvert, &
\lVert g \rVert_\infty &= \max_k \lvert g_k \rvert.
\end{align}
The ratio is invariant under scaling of $f$ and of its domain, so $v = \lVert g \rVert_2^2 / (\lVert g \rVert_1 \lVert g \rVert_\infty)$ is the exact bound of the step function with values $f_i$.
The score uses this recomputed $v$.

\subsection{Third Autocorrelation Inequality}
\label{app:third-autocorrelation}

\paragraph{Problem.}
The third autocorrelation inequality extends the first one to functions that take values of either sign: for every function $f$ supported on $[-\tfrac14, \tfrac14]$,
\begin{align}
\max_{-1/2 \le t \le 1/2} \lvert (f \ast f)(t) \rvert \ge C_3 \left( \int_{-1/4}^{1/4} f(x) \, dx \right)^2,
\end{align}
where $(f \ast f)(t) = \int f(t - x) f(x) \, dx$ is the autoconvolution of $f$.
Every such $f$ with a nonzero integral gives an upper bound on $C_3$.
Find a function $f$ supported on $[-\tfrac14, \tfrac14]$, with values of either sign and a nonzero integral, that minimizes this bound,
\begin{align}
\min_{f} \; \frac{\max_t \lvert (f \ast f)(t) \rvert}{\left( \int f \right)^2}.
\end{align}
The program represents $f$ by its values $f_1, \dots, f_N$ on $N$ equal intervals of $[-\tfrac14, \tfrac14]$ and chooses $N$.

\paragraph{Evaluator.}
The entry function \texttt{run()} returns $f_1, \dots, f_N$, the bound $\tilde v$, a loss value, and $N$.
With the interval width $\Delta = 1/(2N)$, the evaluator computes
\begin{align}
v = \frac{\max_k \Delta \left\lvert \sum_i f_i f_{k-i} \right\rvert}{\left( \Delta \sum_i f_i \right)^2}.
\end{align}
The autoconvolution of a step function is piecewise linear with breakpoints at multiples of $\Delta$, so $v$ is the exact bound of the step function with values $f_i$.
The evaluator requires $(\Delta \sum_i f_i)^2 \ge 10^{-9}$ and $\lvert v - \tilde v \rvert \le 10^{-3}$, and the score uses $\tilde v$.
The evaluator divides by 1.4557, AlphaEvolve's bound on $\lvert \max_t (f \ast f)(t) \rvert / (\int f)^2$; for $v$ as defined here, AlphaEvolve's value is $v_{\mathrm{AE}} = 1.4688$.

\subsection{Uncertainty Inequality}
\label{app:uncertainty}

\paragraph{Problem.}
The uncertainty principle of \citet{bourgain2010principe} bounds from below the product of the radii beyond which an even function and its Fourier transform are non-negative.
For a real even function $f$ with Fourier transform $\hat f(\xi) = \int f(x) e^{-2\pi i x \xi} \, dx$, $f(0) \le 0$, and $\hat f(0) \le 0$, let
\begin{align}
A(f) = \inf \{ r > 0 : f(x) \ge 0 \text{ for all } \lvert x \rvert \ge r \}
\end{align}
be the radius of the last sign change of $f$.
The inequality $A(f) A(\hat f) \ge C_4$ holds for a positive constant $C_4$, so every such $f$ gives an upper bound on $C_4$.
Following \citet{goncalves2017hermite}, the search runs over $f(x) = P(\sqrt{2\pi}\, x) \, e^{-\pi x^2}$ with $P = \sum_{k=0}^{m} a_k H_{4k}$ and $P(0) = 0$, where $H_{4k}$ is the Hermite polynomial of degree $4k$.
Each such $f$ equals its own Fourier transform and satisfies $f(0) = \hat f(0) = 0$, so $A(f) A(\hat f) = r_{\max}^2 / (2\pi)$, where $r_{\max}$ is the largest positive root at which $P$ changes sign.
Find coefficients $a_0, \dots, a_m$ that minimize this bound,
\begin{align}
\min_{a_0, \dots, a_m} \; \frac{r_{\max}^2}{2\pi}.
\end{align}

\paragraph{Evaluator.}
The entry function \texttt{run()} returns $a_0, \dots, a_{m-1}$, the bound $\tilde v$, and the root $\tilde r_{\max}$.
The evaluator builds $P$ in exact rational arithmetic with SymPy: it converts every $a_k$ to a rational number, chooses $a_m = -\sum_{k < m} a_k H_{4k}(0) / H_{4m}(0)$ so that $P(0) = 0$, and multiplies $P$ by $-1$ when $P(x) \to -\infty$ as $x \to \infty$.
It divides $P$ by $x^2$, isolates the real roots of the quotient, evaluates every root to 200 digits, and keeps the roots at which the quotient changes sign.
The largest of these roots is $r_{\max}$, and the objective value is $v = r_{\max}^2 / (2\pi)$.
The evaluator requires $\tilde v$ and $\tilde r_{\max}$ to agree with $v$ and $r_{\max}$ to within a relative and an absolute tolerance of $10^{-9}$, and the score uses the recomputed $v$.
$v_{\mathrm{AE}} = 0.3216$ comes from Laguerre polynomials, outside this family; AlphaEvolve's Hermite construction gives 0.3521.

\subsection{Erd\H{o}s Minimum Overlap Problem}
\label{app:erdos}

\paragraph{Problem.}
For a partition of $\{1, \dots, 2n\}$ into two sets $X$ and $Y$ of $n$ elements each, let $M_k$ be the number of pairs $(x, y) \in X \times Y$ with $x - y = k$.
The minimum overlap problem of Erd\H{o}s asks for the smallest possible value of $\max_k M_k$, and this minimum divided by $n$ converges to a constant $C_5$ as $n \to \infty$~\citep{haugland2016minimum}.
In the continuous formulation, every function $h : [0, 2] \to [0, 1]$ with $\int_0^2 h(x) \, dx = 1$ gives an upper bound on $C_5$.
Find such a function $h$ that minimizes this bound,
\begin{align}
\min_{h} \; \max_k \int h(x) \, \bar h(x + k) \, dx, \qquad \bar h = \begin{cases} 1 - h & \text{on } [0, 2], \\ 0 & \text{elsewhere}. \end{cases}
\end{align}
The program represents $h$ by its values $h_1, \dots, h_N$ on $N$ equal intervals of $[0, 2]$ and chooses $N$.

\paragraph{Evaluator.}
The entry function \texttt{run()} returns $h_1, \dots, h_N$, the bound $\tilde v$, and $N$.
The evaluator requires every $h_i$ to lie in $[0, 1]$ and, with the interval width $\Delta = 2/N$, requires $\lvert \Delta \sum_i h_i - 1 \rvert \le 10^{-3}$.
It computes
\begin{align}
v = \max_k \Delta \sum_i h_i \left( 1 - h_{i+k} \right),
\end{align}
where $k$ ranges over all shifts and the sum runs over the indices with $1 \le i + k \le N$.
It requires $\lvert v - \tilde v \rvert \le 10^{-4}$, and the score uses $\tilde v$.
Our reported scores instead correct $h$ to integrate exactly to 1 and recompute $v$.

\subsection{Sums and Differences of Finite Sets}
\label{app:sums-differences}

\paragraph{Problem.}
For a finite set $U$ of integers, \citet{gyarmati2007sums} study how large the difference set $U - U = \{ u - u' : u, u' \in U \}$ can be relative to the sum set $U + U = \{ u + u' : u, u' \in U \}$.
For a finite set $U$ of non-negative integers with $0 \in U$, the quantity
\begin{align}
1 + \frac{\log \left( \lvert U - U \rvert / \lvert U + U \rvert \right)}{\log \left( 2 \max U + 1 \right)}
\end{align}
is a lower bound on the constant $C_6$ of this problem.
Find a finite set $U$ of non-negative integers with $0 \in U$ that maximizes this bound.

\paragraph{Evaluator.}
The entry function \texttt{run()} returns $U$ as a one-dimensional array and the bound $\tilde v$.
The evaluator requires $0 \in U$ and every element of $U$ to be a non-negative integer.
It computes $U + U$ and $U - U$ from all pairwise sums and differences, recomputes the objective value $v$, requires $v$ and $\tilde v$ to agree to within a relative tolerance of $10^{-5}$ and an absolute tolerance of $10^{-8}$, and the score uses $\tilde v$.

\subsection{Kissing Number in Dimension 11}
\label{app:kissing}

\noindent
\begin{minipage}[t]{0.63\textwidth}
\vspace{0pt}
\paragraph{Problem.}
The kissing number $\tau_{11}$ is the largest number of unit balls in $\mathbb{R}^{11}$ with disjoint interiors that touch a central unit ball~\citep{conway1999sphere}.
A set $S \subset \mathbb{Z}^{11} \setminus \{0\}$ whose largest norm is at most its smallest pairwise distance satisfies $\langle p, q \rangle \le \tfrac12 \lVert p \rVert \lVert q \rVert$ for every pair, so the unit balls centered at the points $2p / \lVert p \rVert$ touch the central unit ball and have disjoint interiors, which proves $\tau_{11} \ge \lvert S \rvert$; \autoref{fig:kissing} shows the two-dimensional analogue.
Find the largest such set:
\begin{align}
\max_{S \subset \mathbb{Z}^{11} \setminus \{0\}} \quad & \lvert S \rvert \nonumber \\
\text{subject to} \quad & \max_{p \in S} \lVert p \rVert \le \min_{p, q \in S,\, p \neq q} \lVert p - q \rVert.
\end{align}
\end{minipage}\hfill
\begin{minipage}[t]{0.33\textwidth}
\vspace{0pt}
\centering
\begin{tikzpicture}[x=0.5cm, y=0.5cm]
\draw[brown, thick, fill=brown!10] (0,0) circle[radius=1];
\fill[brown] (0,0) circle[radius=1.2pt];
\draw[plotBlue, fill=plotBlue!15] (2.0000,0.0000) circle[radius=1];
\draw[brown!60, -{Stealth[length=3pt]}] (0,0) -- (2.0000,0.0000);
\node[font=\footnotesize, text=plotBlue] at (4.0000,0.0000) {$(10, 0)$};
\draw[plotBlue, fill=plotBlue!15] (0.5747,1.9157) circle[radius=1];
\draw[brown!60, -{Stealth[length=3pt]}] (0,0) -- (0.5747,1.9157);
\node[font=\footnotesize, text=plotBlue] at (1.1494,3.8313) {$(3, 10)$};
\draw[plotBlue, fill=plotBlue!15] (-1.6000,1.2000) circle[radius=1];
\draw[brown!60, -{Stealth[length=3pt]}] (0,0) -- (-1.6000,1.2000);
\node[font=\footnotesize, text=plotBlue] at (-3.2000,2.4000) {$(-8, 6)$};
\draw[plotBlue, fill=plotBlue!15] (-1.6000,-1.2000) circle[radius=1];
\draw[brown!60, -{Stealth[length=3pt]}] (0,0) -- (-1.6000,-1.2000);
\node[font=\footnotesize, text=plotBlue] at (-3.2000,-2.4000) {$(-8, -6)$};
\draw[plotBlue, fill=plotBlue!15] (0.5747,-1.9157) circle[radius=1];
\draw[brown!60, -{Stealth[length=3pt]}] (0,0) -- (0.5747,-1.9157);
\node[font=\footnotesize, text=plotBlue] at (1.1494,-3.8313) {$(3, -10)$};
\end{tikzpicture}
\captionsetup{type=figure}\caption{The two-dimensional analogue: five points $p$ of $\mathbb{Z}^2$ that satisfy the constraint, and the unit circles centered at $2p / \lVert p \rVert$, which touch the central unit circle.}
\label{fig:kissing}
\end{minipage}

\paragraph{Evaluator.}
The entry function \texttt{kissing\_number11()} returns an array with 11 columns.
The evaluator rounds every coordinate to the nearest integer and computes the squared norms and the squared pairwise distances in exact integer arithmetic.
It requires the smallest squared norm to be positive and the smallest squared pairwise distance to be at least the largest squared norm, and the objective value is $v = \lvert S \rvert$.

\subsection{Sparse Linear Systems}
\label{app:sparse-linear-systems}

\paragraph{Problem.}
Solve $A x = b$ for five sparse matrices $A$ from finite-difference discretizations of partial differential equations on the unit square or cube with Dirichlet boundary conditions~\citep{saad2003iterative}:
\begin{enumerate}[leftmargin=*]
\item the two-dimensional Poisson equation with the 5-point stencil on a $256 \times 256$ grid ($n = 65{,}536$ unknowns), symmetric positive definite;
\item the three-dimensional diffusion equation $-\nabla \cdot (c \nabla u) = f$ with the 7-point stencil on a $64^3$ grid ($n = 262{,}144$), where $c$ is constant on the blocks of an $8 \times 8 \times 8$ partition with values $10^{s}$, $s$ uniform in $[0, 6]$, and harmonic means on the cell faces, symmetric positive definite and ill-conditioned;
\item the two-dimensional Helmholtz operator $-\Delta - k^2$ with $k = 40.3$ and the 5-point stencil on a $300 \times 300$ grid ($n = 90{,}000$), symmetric indefinite;
\item the two-dimensional convection-diffusion operator $-\varepsilon \Delta u + w \cdot \nabla u$ with the rotating flow $w = (2y - 1, 1 - 2x)$, $\varepsilon = 10^{-3}$, and first-order upwind differences on a $300 \times 300$ grid ($n = 90{,}000$), nonsymmetric;
\item the three-dimensional convection-diffusion operator with the flow $w = (2y - 1, 1 - 2x, 0.5)$, $\varepsilon = 10^{-2}$, and first-order upwind differences on a $48^3$ grid ($n = 110{,}592$), nonsymmetric.
\end{enumerate}

\paragraph{Evaluator.}
The entry function \texttt{solve(A, b)} receives $A$ as a SciPy sparse matrix in compressed sparse row format and returns $x$.
Each right-hand side is $b = A x^\star$ for a solution $x^\star$ with independent standard normal entries, and the relative error is $e = \lVert x - x^\star \rVert_2 / \lVert x^\star \rVert_2$.
The cap is $d_{\mathrm{cap}} = 8$ digits, and the time limit is 9 seconds per case.
The initial program uses the conjugate gradient method with a Jacobi preconditioner and a relative tolerance of $10^{-10}$.

\subsection{Eigenvalues of Non-Normal Matrices}
\label{app:eigenvalues}

\paragraph{Problem.}
Compute all eigenvalues of six square matrices with small integer entries to high relative accuracy.
Backward-stable eigenvalue algorithms return the eigenvalues of a nearby matrix, and the eigenvalues of a non-normal matrix can be far from those of a nearby matrix, so these algorithms lose digits on non-normal matrices~\citep{trefethen2005pseudospectra}.
The six matrices are
\begin{enumerate}[leftmargin=*]
\item a dense circulant matrix of size $n = 200$, which is normal;
\item a tridiagonal Toeplitz matrix with diagonal 9, sub-diagonal 1, and super-diagonal 4, of size $n = 200$;
\item the Kronecker sum of two nonsymmetric tridiagonal Toeplitz matrices of sizes 24 and 22, an upwind discretization of a two-dimensional convection-diffusion operator, of size $n = 528$;
\item a nonsymmetric tridiagonal matrix with random diagonal entries in $\{-12, \dots, 12\}$, sub-diagonal entries in $\{1, 2\}$, and super-diagonal entries in $\{4, \dots, 8\}$, with rows and columns permuted by the same random permutation, of size $n = 200$;
\item a block upper-triangular matrix with $2 \times 2$ diagonal blocks and sparse random entries above them, with rows and columns permuted by the same random permutation, of size $n = 200$;
\item a dense matrix $U B U^{-1}$, where $B$ is block diagonal with $2 \times 2$ blocks and $U$ is a product of elementary integer matrices with determinant 1, of size $n = 180$.
\end{enumerate}
No eigenvalue is zero, and the problem statement names the structure of each matrix.

\paragraph{Evaluator.}
The entry function \texttt{eigenvalues(A)} receives an integer array and returns the $n$ eigenvalues as a complex array in any order.
The evaluator knows every eigenvalue in closed form or from a well-conditioned computation: the discrete Fourier transform of the first row of the circulant matrix, $a + 2\sqrt{bc}\cos(j\pi/(m+1))$ for a tridiagonal Toeplitz matrix of size $m$ with diagonal $a$ and off-diagonals $b$ and $c$, sums of these values for the Kronecker sum, the eigenvalues of the similar symmetric tridiagonal matrix with off-diagonals $\sqrt{b_i c_i}$ for the fourth matrix, and the eigenvalues of the $2 \times 2$ blocks for the last two matrices.
It pairs the returned eigenvalues one-to-one with the true eigenvalues $\lambda_j$ by the assignment that minimizes the total relative error, and the error $e$ of a matrix is the largest relative error $\lvert \mu_j - \lambda_j \rvert / \lvert \lambda_j \rvert$ over the pairs.
The cap is $d_{\mathrm{cap}} = 10$ digits, and the time limit is 8 seconds per case.
The initial program calls the LAPACK eigenvalue routine of NumPy.

\subsection{Viscous Burgers Equation}
\label{app:burgers}

\noindent
\begin{minipage}[t]{0.55\textwidth}
\vspace{0pt}
\paragraph{Problem.}
Solve the one-dimensional viscous Burgers equation
\begin{align}
u_t + u u_x &= \nu u_{xx}, \quad x \in [-1, 1], \quad u(x, 0) = -\sin(\pi x),
\end{align}
with periodic boundary conditions up to the final time $T = 1$, for the five viscosities $\nu \in \{0.1, 0.03, 0.01, 0.003, 0.001\}$.
From $t = 1/\pi$ on, a shock of width proportional to $\nu$ forms at $x = 0$ (\autoref{fig:burgers}).
\end{minipage}\hfill
\begin{minipage}[t]{0.42\textwidth}
\vspace{0pt}
\centering
\input{figures/problem_illustrations/burgers_1d}
\captionsetup{type=figure}\caption{Reference solution $u(x, 1)$ for three viscosities. The shock at $x = 0$ steepens as $\nu$ decreases.}
\label{fig:burgers}
\end{minipage}

\paragraph{Evaluator.}
The entry function \texttt{solve(nu, T, x)} receives 1000 points $x$ drawn uniformly from $[-1, 1]$ and returns $u(x, T)$ at these points.
The reference solution is the Cole--Hopf solution~\citep{hopf1950burgers,cole1951quasilinear},
\begin{align}
u(x, t) &= -\frac{\int \sin(\pi(x - y))\, F(x - y)\, G(y) \, dy}{\int F(x - y)\, G(y) \, dy}, & F(y) &= e^{-\cos(\pi y)/(2\pi\nu)}, \quad G(y) = e^{-y^2/(4\nu t)},
\end{align}
which the evaluator integrates with the trapezoidal rule on 8001 nodes in scaled logarithmic form; four times as many nodes and arbitrary-precision quadrature give the same values.
The error $e$ is the relative 2-norm error over the 1000 points, the cap is $d_{\mathrm{cap}} = 6$ digits, and the time limit is 9 seconds per case.
The initial program uses explicit central finite differences on 256 grid points with forward Euler time steps.

\subsection{Convection-Diffusion Equation}
\label{app:convection-diffusion}

\noindent
\begin{minipage}[t]{0.55\textwidth}
\vspace{0pt}
\paragraph{Problem.}
Solve the steady convection-diffusion equation
\begin{align}
-\varepsilon (u_{xx} + u_{yy}) + b_1 u_x + b_2 u_y &= 1 \quad \text{in } (0, 1)^2, \\ u &= 0 \quad \text{on the boundary},
\end{align}
with the constant flow $b = (1, 0.5)$, for the five diffusion coefficients $\varepsilon \in \{10^{-1}, 10^{-2}, 3 \cdot 10^{-3}, 10^{-3}, 3 \cdot 10^{-4}\}$.
The solution has boundary layers of width proportional to $\varepsilon$ along the outflow sides $x = 1$ and $y = 1$ (\autoref{fig:convection-diffusion}).
\end{minipage}\hfill
\begin{minipage}[t]{0.42\textwidth}
\vspace{0pt}
\centering
\begin{tikzpicture}
\begin{axis}[width=4.6cm, height=4.6cm, scale only axis=false, axis on top, enlargelimits=false,
  xmin=0, xmax=1, ymin=0, ymax=1, xtick={0,1}, ytick={0,1}, xlabel={$x$}, ylabel={$y$}, ylabel style={yshift=-8pt}, xlabel style={yshift=6pt},
  colormap={convection_diffusion_2d}{rgb255=(255,255,255), rgb255=(100,143,255), rgb255=(120,94,240)}, colorbar, point meta min=0, point meta max=0.94,
  colorbar style={width=0.15cm, ytick={0,0.94}, yticklabel style={font=\footnotesize}, title={$u$}, title style={font=\footnotesize}},
  tick label style={font=\footnotesize}, label style={font=\small}]
\addplot graphics[xmin=0, xmax=1, ymin=0, ymax=1] {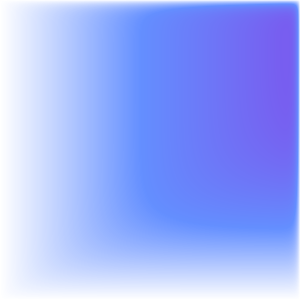};
\end{axis}
\end{tikzpicture}
\captionsetup{type=figure}\caption{Reference solution for $\varepsilon = 10^{-2}$, with boundary layers along $x = 1$ and $y = 1$.}
\label{fig:convection-diffusion}
\end{minipage}

\paragraph{Evaluator.}
The entry function \texttt{solve(eps, b, x, y)} receives 2000 points drawn uniformly from the unit square and returns $u$ at these points.
The reference solution comes from Chebyshev collocation~\citep{trefethen2000spectral}: with constant coefficients, the collocation equations on the interior points form a Sylvester equation, which the evaluator solves with the Bartels--Stewart algorithm~\citep{bartels1972sylvester} and interpolates to the points with barycentric interpolation.
Each reference uses two resolutions, between 60 and 1200 Chebyshev points per direction depending on $\varepsilon$, that agree to $10^{-8}$.
The error $e$ is the relative 2-norm error over the 2000 points, the cap is $d_{\mathrm{cap}} = 6$ digits, and the time limit is 9 seconds per case.
The initial program uses central finite differences on a $64 \times 64$ grid with a sparse direct solver and bilinear interpolation.

\subsection{Helmholtz Equation}
\label{app:helmholtz}

\noindent
\begin{minipage}[t]{0.55\textwidth}
\vspace{0pt}
\paragraph{Problem.}
Solve the Helmholtz equation with the impedance boundary condition
\begin{align}
-(u_{xx} + u_{yy}) - k^2 u &= f \quad \text{in } (0, 1)^2, \\ \partial_n u - i k u &= 0 \quad \text{on the boundary},
\end{align}
where $\partial_n$ is the outward normal derivative, for the five wave numbers $k \in \{10, 20, 40, 60, 80\}$.
The source consists of two Gaussians whose width is proportional to the wavelength,
\begin{align}
f(x, y) &= e^{-\frac{k^2}{2}\left((x - 0.3)^2 + (y - 0.6)^2\right)} \nonumber \\ &\quad - 0.6\, e^{-\frac{k^2}{2}\left((x - 0.7)^2 + (y - 0.25)^2\right)},
\end{align}
and the solution $u$ is complex (\autoref{fig:helmholtz}).
At a fixed number of grid points per wavelength, the error of low-order discretizations grows with $k$~\citep{babuska1997pollution}.
\end{minipage}\hfill
\begin{minipage}[t]{0.42\textwidth}
\vspace{0pt}
\centering
\begin{tikzpicture}
\begin{axis}[width=4.6cm, height=4.6cm, scale only axis=false, axis on top, enlargelimits=false,
  xmin=0, xmax=1, ymin=0, ymax=1, xtick={0,1}, ytick={0,1}, xlabel={$x$}, ylabel={$y$}, ylabel style={yshift=-8pt}, xlabel style={yshift=6pt},
  colormap={helmholtz_2d}{rgb255=(100,143,255), rgb255=(255,255,255), rgb255=(254,97,0)}, colorbar, point meta min=-3.6, point meta max=3.6,
  colorbar style={width=0.15cm, ytick={-3.6,3.6}, yticklabel style={font=\footnotesize}, title={$10^4\, \mathrm{Re}\, u$}, title style={font=\footnotesize}},
  tick label style={font=\footnotesize}, label style={font=\small}]
\addplot graphics[xmin=0, xmax=1, ymin=0, ymax=1] {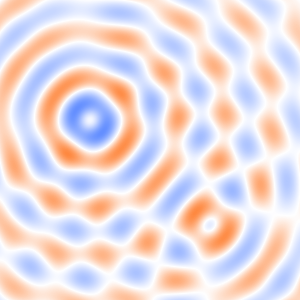};
\end{axis}
\end{tikzpicture}
\captionsetup{type=figure}\caption{Real part of the reference solution for $k = 40$, with waves radiating from the two sources.}
\label{fig:helmholtz}
\end{minipage}

\paragraph{Evaluator.}
The entry function \texttt{solve(k, x, y)} receives 2000 points drawn uniformly from the unit square and returns the complex values of $u$ at these points.
The reference solution comes from Chebyshev collocation with the boundary values eliminated through the impedance condition in each direction, which again gives a Sylvester equation~\citep{trefethen2000spectral,bartels1972sylvester}.
Each reference uses two resolutions, between 60 and 320 Chebyshev points per direction depending on $k$, that agree to $10^{-8}$, and second-order finite differences converge to it at the expected rate.
The error $e$ is the relative 2-norm error over the 2000 points, the cap is $d_{\mathrm{cap}} = 6$ digits, and the time limit is 9 seconds per case.
The initial program uses second-order finite differences on a $101 \times 101$ grid, with ghost points for the impedance condition, and a sparse direct solver.

\section{Best Known Results}
\label{app:best-known}

Supplementary Table~\ref{tab:alphaevolve-raw} lists the objective values behind the scores of Table~\ref{tab:results-60min}: the value of AlphaEvolve's published construction, the best value of the three runs of each framework, and the best known value. We rerun the best program of every run and recompute its value under AlphaEvolve's exact constraints. For the programs that reproduce their logged score, the exact constraints lower the score by at most $3.3 \times 10^{-4}$. The remaining programs draw unseeded random numbers, so we rerun each of them 33 times (once without a seed and with seeds 0 to 31) and report the best value under the exact constraints among the reruns that the evaluator scores no higher than the logged score of the run. We call a best known value verified if its construction or certificate has been checked, and claimed otherwise.

With Qwen3 8B and 32B-FP8, SwarmEvolve improves on AlphaEvolve on the same three problems: it matches the best known packing of 26 circles in the unit square~\citep{specht2026packomania}, and its step functions improve AlphaEvolve's bounds on the second and third autocorrelation inequalities. The three baselines improve on AlphaEvolve only on the autocorrelation inequalities.

Two problems need care in the comparison with AlphaEvolve. On the third autocorrelation inequality, the evaluator of OpenEvolve takes the absolute value of the autoconvolution inside the maximum, while AlphaEvolve's published bound of 1.4557 takes it outside the maximum. Under the definition of the evaluator, AlphaEvolve's construction gives 1.4688, the value that we compare against (Appendix~\ref{app:third-autocorrelation}). On the uncertainty inequality, the evaluator accepts only functions built from Hermite polynomials. Within this family, AlphaEvolve's first construction gives 0.3521, while its best bound of 0.3216 uses Laguerre polynomials, which lie outside the family. SwarmEvolve reaches 0.3521 with Qwen3 32B-FP8, the value of AlphaEvolve's construction within the same family.

With Qwen3 32B-FP8, SwarmEvolve comes close to AlphaEvolve on four more problems: within one part per million on circles in a rectangle, and within 0.02\% on the two distance ratios and the minimum overlap problem. For all four frameworks, the largest gaps to AlphaEvolve include the hexagon packings, the Heilbronn problems, and the kissing number in dimension 11.

\begin{table}[H]
\centering
\caption{Best values on the 16 AlphaEvolve problems: the value of AlphaEvolve, the best of three 60-minute runs of each framework with Qwen3 8B, 14B, and 32B-FP8, and the best known value to the best of our knowledge. Values are given to seven significant digits. Bold: best of all values of a problem; underlined: second best; $^\dagger$: better than AlphaEvolve; --: no valid program within 60 minutes.}
\label{tab:alphaevolve-raw}
\footnotesize
\setlength{\tabcolsep}{3pt}
\renewcommand{\arraystretch}{1.15}
\begin{adjustbox}{max width=\textwidth}
\begin{tabular}{llll>{\columncolor{oursShade}}llllll}
\toprule
\color{brown}\textbf{Problem} & \color{brown}\textbf{Objective} & AlphaEvolve & \color{brown}\textbf{Qwen3} & \textbf{Ours} & OpenEvolve & ShinkaEvolve & CodeEvolve & \makecell[l]{Best known\\(Nov.~5, 2026)} & Status \\
\midrule
 &  &  & 8B & \best{2.635983}$^\dagger$ & 2.615077 & 2.626003 & 2.613583 &  &  \\
 &  &  & 14B & 2.634292 & 2.346719 & 2.630328 & 2.616260 &  &  \\
\multirow{-3}{*}{Circles in a square} & \multirow{-3}{*}{$\sum_i r_i$ $\uparrow$} & \multirow{-3}{*}{\second{2.635863}} & 32B-FP8 & \best{2.635983}$^\dagger$ & 2.613827 & 2.629864 & 2.620510 & \multirow{-3}{*}{\best{2.635983}} & \multirow{-3}{*}{verified~\citep{specht2026packomania}} \\
\rowcolor{rowShade} &  &  & 8B & \cellcolor{oursShade}2.100000 & 2.332669 & 2.357264 & 2.342198 &  &  \\
\rowcolor{rowShade} &  &  & 14B & \cellcolor{oursShade}2.357262 & 2.268246 & 2.295059 & 2.357264 &  &  \\
\rowcolor{rowShade}\multirow{-3}{*}{Circles in a rectangle} & \multirow{-3}{*}{$\sum_i r_i$ $\uparrow$} & \multirow{-3}{*}{\best{2.365832}} & 32B-FP8 & \cellcolor{oursShade}\second{2.365830} & 2.313769 & 2.272727 & 2.346834 & \multirow{-3}{*}{\best{2.365832}} & \multirow{-3}{*}{verified~\citep{berthold2026global}} \\
 &  &  & 8B & 4.505917 & 6.002853 & 8.000000 & 8.000000 &  &  \\
 &  &  & 14B & 4.677289 & 5.404701 & 6.031025 & 6.333333 &  &  \\
\multirow{-3}{*}{Hexagons, $n=11$} & \multirow{-3}{*}{$R$ $\downarrow$} & \multirow{-3}{*}{\second{3.930092}} & 32B-FP8 & 4.472263 & 5.004000 & 5.418582 & 8.000000 & \multirow{-3}{*}{\best{3.924501}} & \multirow{-3}{*}{verified~\citep{friedman2026hexinhex}} \\
\rowcolor{rowShade} &  &  & 8B & \cellcolor{oursShade}4.924979 & 6.031025 & 6.031025 & 8.000000 &  &  \\
\rowcolor{rowShade} &  &  & 14B & \cellcolor{oursShade}4.537633 & 5.080000 & 6.031025 & 5.595754 &  &  \\
\rowcolor{rowShade}\multirow{-3}{*}{Hexagons, $n=12$} & \multirow{-3}{*}{$R$ $\downarrow$} & \multirow{-3}{*}{\second{3.941912}} & 32B-FP8 & \cellcolor{oursShade}5.029604 & 5.000400 & 6.829775 & 8.000000 & \multirow{-3}{*}{\best{3.94164}} & \multirow{-3}{*}{claimed~\citep{friedman2026hexinhex}} \\
\midrule
 &  &  & 8B & 0.02227742 & 0.001599633 & 0.02003859 & 0.02217071 &  &  \\
 &  &  & 14B & 0.02600437 & 0.009468544 & 0.01510912 & 0.02590121 &  &  \\
\multirow{-3}{*}{Heilbronn, triangle} & \multirow{-3}{*}{$A_{\min} / A$ $\uparrow$} & \multirow{-3}{*}{\best{0.03652989}} & 32B-FP8 & \second{0.03064911} & 0.003508768 & 0.02927480 & 0.02144022 & \multirow{-3}{*}{\best{0.03652989}} & \multirow{-3}{*}{verified~\citep{novikov2025alphaevolve}} \\
\rowcolor{rowShade} &  &  & 8B & \cellcolor{oursShade}0.02302219 & 0.02052219 & 0.02195328 & 0.02002288 &  &  \\
\rowcolor{rowShade} &  &  & 14B & \cellcolor{oursShade}0.02341767 & 0.02202470 & 0.02224608 & 0.02009201 &  &  \\
\rowcolor{rowShade}\multirow{-3}{*}{Heilbronn, convex, $n=13$} & \multirow{-3}{*}{$A_{\min} / A_{\mathrm{hull}}$ $\uparrow$} & \multirow{-3}{*}{\second{0.03093689}} & 32B-FP8 & \cellcolor{oursShade}0.02769717 & 0.01762215 & 0.02230236 & 0.02145342 & \multirow{-3}{*}{\best{0.03093721}} & \multirow{-3}{*}{verified~\citep{tejstead2026heilbronn}} \\
 &  &  & 8B & 0.02261399 & 0.01414730 & 0.01414730 & 0.01414730 &  &  \\
 &  &  & 14B & 0.02229207 & 0.01807289 & 0.01980031 & 0.01762215 &  &  \\
\multirow{-3}{*}{Heilbronn, convex, $n=14$} & \multirow{-3}{*}{$A_{\min} / A_{\mathrm{hull}}$ $\uparrow$} & \multirow{-3}{*}{\second{0.02783557}} & 32B-FP8 & 0.02232631 & 0.01893828 & 0.01980032 & 0.01807289 & \multirow{-3}{*}{\best{0.02783558}} & \multirow{-3}{*}{verified~\citep{tejstead2026heilbronn}} \\
\rowcolor{rowShade} &  &  & 8B & \cellcolor{oursShade}12.94885 & 18.00000 & 13.20266 & 13.15947 &  &  \\
\rowcolor{rowShade} &  &  & 14B & \cellcolor{oursShade}12.89897 & 18.00000 & 13.01783 & 14.20428 &  &  \\
\rowcolor{rowShade}\multirow{-3}{*}{Distance ratio, plane} & \multirow{-3}{*}{$(d_{\max} / d_{\min})^2$ $\downarrow$} & \multirow{-3}{*}{\second{12.88927}} & 32B-FP8 & \cellcolor{oursShade}12.88956 & 18.00000 & 13.27883 & 17.01391 & \multirow{-3}{*}{\best{12.88923}} & \multirow{-3}{*}{verified~\citep{einsteinarena2026}} \\
 &  &  & 8B & 4.177907 & 4.732051 & 4.546903 & 4.474247 &  &  \\
 &  &  & 14B & 4.167784 & 4.732051 & 4.586633 & 4.352043 &  &  \\
\multirow{-3}{*}{Distance ratio, space} & \multirow{-3}{*}{$(d_{\max} / d_{\min})^2$ $\downarrow$} & \multirow{-3}{*}{\second{4.165850}} & 32B-FP8 & 4.166486 & 4.540188 & 4.370018 & 4.630033 & \multirow{-3}{*}{\best{4.165784}} & \multirow{-3}{*}{verified~\citep{berthold2026global}} \\
\midrule
\rowcolor{rowShade} &  &  & 8B & \cellcolor{oursShade}1.512329 & 1.515894 & 1.516525 & 1.516662 &  &  \\
\rowcolor{rowShade} &  &  & 14B & \cellcolor{oursShade}1.512125 & 1.515684 & 1.515220 & 1.517188 &  &  \\
\rowcolor{rowShade}\multirow{-3}{*}{Autocorrelation I} & \multirow{-3}{*}{$C_1$ $\downarrow$} & \multirow{-3}{*}{\second{1.505294}} & 32B-FP8 & \cellcolor{oursShade}1.512209 & 1.516507 & 1.510459 & 1.516020 & \multirow{-3}{*}{\best{1.502743}} & \multirow{-3}{*}{verified~\citep{tao2026optimizationproblems}} \\
 &  &  & 8B & 0.9213998$^\dagger$ & 0.9034590$^\dagger$ & 0.8995052$^\dagger$ & 0.9032550$^\dagger$ &  &  \\
 &  &  & 14B & \second{0.9314238}$^\dagger$ & 0.9198818$^\dagger$ & 0.9064627$^\dagger$ & 0.9001606$^\dagger$ &  &  \\
\multirow{-3}{*}{Autocorrelation II} & \multirow{-3}{*}{$C_2$ $\uparrow$} & \multirow{-3}{*}{0.8962799} & 32B-FP8 & 0.9289826$^\dagger$ & 0.8879468 & 0.9042086$^\dagger$ & 0.8880995 & \multirow{-3}{*}{\best{0.9635881}} & \multirow{-3}{*}{verified~\citep{einsteinarena2026}} \\
\rowcolor{rowShade} &  &  & 8B & \cellcolor{oursShade}1.460021$^\dagger$ & 1.513955 & 1.478389 & 1.477685 &  &  \\
\rowcolor{rowShade} &  &  & 14B & \cellcolor{oursShade}1.460180$^\dagger$ & 1.473397 & 1.468036$^\dagger$ & 1.476167 &  &  \\
\rowcolor{rowShade}\multirow{-3}{*}{Autocorrelation III} & \multirow{-3}{*}{$C_3$ $\downarrow$} & \multirow{-3}{*}{1.468762} & 32B-FP8 & \cellcolor{oursShade}\second{1.459198}$^\dagger$ & 1.469457 & 1.468273$^\dagger$ & 1.478609 & \multirow{-3}{*}{\best{1.453675}} & \multirow{-3}{*}{verified~\citep{ye2026simpletes}} \\
 &  &  & 8B & 0.3524471 & 0.3566070 & -- & 0.3615746 &  &  \\
 &  &  & 14B & 0.3525058 & 0.3560237 & -- & 0.3531502 &  &  \\
\multirow{-3}{*}{Uncertainty} & \multirow{-3}{*}{$C_4$ $\downarrow$} & \multirow{-3}{*}{\second{0.3215872}} & 32B-FP8 & 0.3521054 & 0.3522484 & -- & 0.3585374 & \multirow{-3}{*}{\best{0.3089}} & \multirow{-3}{*}{verified~\citep{chung2026station}} \\
\midrule
\rowcolor{rowShade} &  &  & 8B & \cellcolor{oursShade}0.3810116 & 0.3816114 & 0.3815447 & 0.3834166 &  &  \\
\rowcolor{rowShade} &  &  & 14B & \cellcolor{oursShade}0.3810021 & 0.3810436 & 0.3810753 & 0.3816378 &  &  \\
\rowcolor{rowShade}\multirow{-3}{*}{Erd\H{o}s overlap} & \multirow{-3}{*}{$C_5$ $\downarrow$} & \multirow{-3}{*}{\second{0.3809230}} & 32B-FP8 & \cellcolor{oursShade}0.3809729 & 0.3812057 & 0.3810462 & 0.3814869 & \multirow{-3}{*}{\best{0.3808563}} & \multirow{-3}{*}{verified~\citep{einsteinarena2026}} \\
 &  &  & 8B & 1.109543 & 1.097861 & -- & 1.103907 &  &  \\
 &  &  & 14B & 1.109543 & 1.101635 & 1.037493 & 1.090860 &  &  \\
\multirow{-3}{*}{Sums and differences} & \multirow{-3}{*}{$C_6$ $\uparrow$} & \multirow{-3}{*}{\second{1.158417}} & 32B-FP8 & 1.122610 & 1.103165 & 1.011447 & 1.107311 & \multirow{-3}{*}{\best{1.193597}} & \multirow{-3}{*}{claimed~\citep{tao2026optimizationproblems}} \\
\rowcolor{rowShade} &  &  & 8B & \cellcolor{oursShade}220 & 416 & 416 & 220 &  &  \\
\rowcolor{rowShade} &  &  & 14B & \cellcolor{oursShade}416 & 464 & 480 & 220 &  &  \\
\rowcolor{rowShade}\multirow{-3}{*}{Kissing number, $d=11$} & \multirow{-3}{*}{$\lvert S \rvert$ $\uparrow$} & \multirow{-3}{*}{\second{593}} & 32B-FP8 & \cellcolor{oursShade}476 & 464 & 528 & 220 & \multirow{-3}{*}{\best{604}} & \multirow{-3}{*}{verified~\citep{bianchi2026collective}} \\
\bottomrule
\end{tabular}
\end{adjustbox}
\end{table}

\section{Best Solutions Found by Each Framework}
\label{app:best-solutions}

We compare the best program that each framework found with Qwen3 8B for one packing problem, one geometry problem, and the three PDE problems.
For each framework, the highest-scoring program is selected.
For the PDE problems, we plot the pointwise absolute error $|u - u_{\mathrm{ref}}|$, where $u$ is each framework's solution and $u_{\mathrm{ref}}$ is the reference solution, on the same logarithmic scale for all three problems.

\subsection{Circle Packing in the Unit Square}

\begin{figure}[H]
\centering
\includegraphics[width=\textwidth]{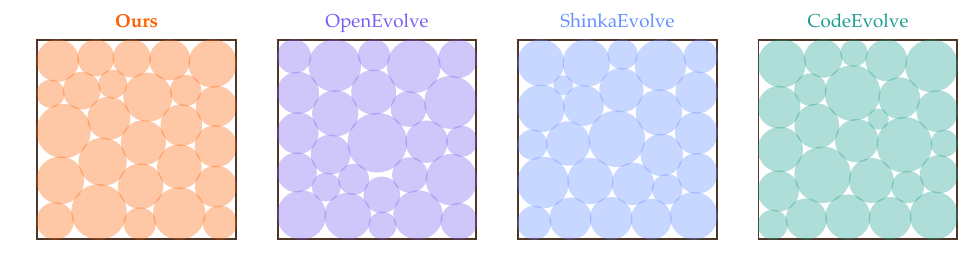}
\caption{Best packings of 26 circles in the unit square. The sums of the radii are 2.636 (our method), 2.615 (OpenEvolve), 2.626 (ShinkaEvolve), and 2.614 (CodeEvolve).}
\label{fig:best-circle-packing}
\end{figure}

\autoref{fig:best-circle-packing} shows the best packing of each framework.

\paragraph{Our method.}
The program runs ten restarts of one SLSQP optimization over all centers and radii, each from 16 perturbed lattice positions and 10 random positions, and keeps the best valid packing.
\paragraph{OpenEvolve.}
The program places one circle at the center, eight on an inner ring, four in the corners, and thirteen on an outer ring, and optimizes all centers and radii with one SLSQP run.
\paragraph{ShinkaEvolve.}
The program places one circle at the center, eight on an inner ring, thirteen on an outer ring clipped to the square, and four near the corners, and optimizes all centers and radii with one SLSQP run.
\paragraph{CodeEvolve.}
The program places 25 circles on a $5 \times 5$ lattice and one at the center, and optimizes all centers and radii with one SLSQP run.

\subsection{Heilbronn Problem in a Triangle}

\begin{figure}[H]
\centering
\includegraphics[width=\textwidth]{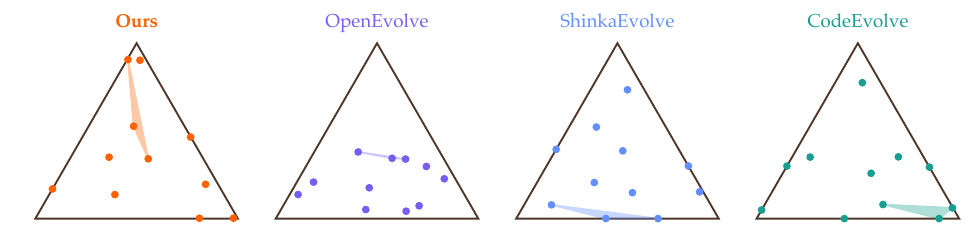}
\caption{Best placements of 11 points in the equilateral triangle. The shaded triangle has the smallest area among all triangles formed by the points. The smallest areas, divided by the area of the triangle, are 0.0223 (our method), 0.0016 (OpenEvolve), 0.0200 (ShinkaEvolve), and 0.0222 (CodeEvolve).}
\label{fig:best-heilbronn-triangle}
\end{figure}

\autoref{fig:best-heilbronn-triangle} shows the best placement of each framework.

\paragraph{Our method.}
The program runs ten restarts of simulated annealing from a symmetric lattice of points, moving one point at a time and accepting a worse smallest area with a probability that decreases as the temperature drops.
\paragraph{OpenEvolve.}
The program draws 11 random points in the triangle, redrawing a point whenever it is collinear with two earlier points.
\paragraph{ShinkaEvolve.}
The program draws random configurations from 50 fixed seeds, keeps the best ones, and refines them with SLSQP on the smallest triangle area under the three edge constraints of the triangle.
\paragraph{CodeEvolve.}
The program draws 11 random points in the triangle and runs one simulated annealing of 50{,}000 steps that moves one point at a time.

\subsection{Viscous Burgers Equation}

\begin{figure}[H]
\centering
\includegraphics[width=\textwidth]{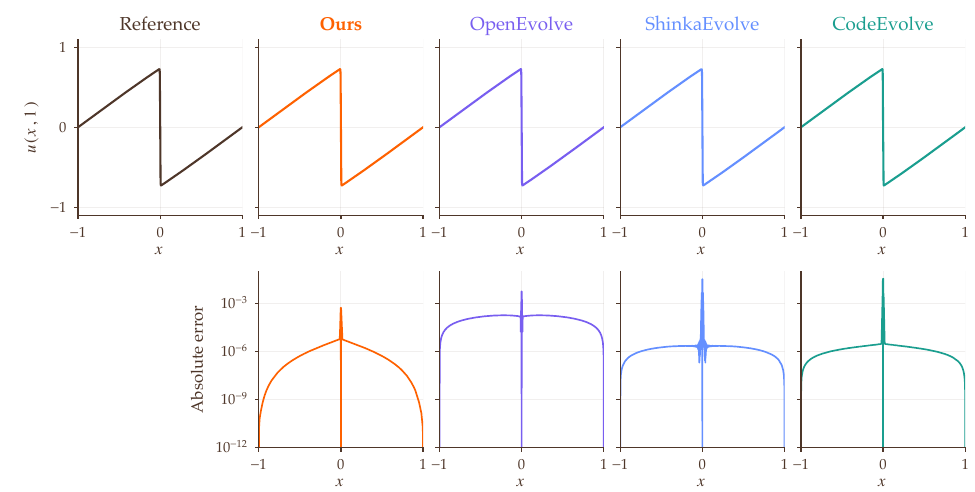}
\caption{Solutions $u(x, 1)$ of the Burgers equation for $\nu = 0.001$ (top) and their absolute errors (bottom). Our solution is the most accurate away from the shock at $x = 0$.}
\label{fig:best-burgers}
\end{figure}

\autoref{fig:best-burgers} shows the solution of each framework for the smallest viscosity and its error.

\paragraph{Our method.}
The program uses central finite differences on 8192 grid points for $\nu < 0.01$ and 2048 points otherwise, with a two-stage Runge--Kutta method and a time step limited by both the advection and the diffusion stability conditions.
\paragraph{OpenEvolve.}
The program uses central finite differences of the flux $u^2/2$ on 2048 grid points with forward Euler time steps.
\paragraph{ShinkaEvolve.}
The program splits each time step into a two-stage Runge--Kutta step for the advection on 1024 grid points and an implicit diffusion step solved exactly in Fourier space.
\paragraph{CodeEvolve.}
The program uses central finite differences of the flux $u^2/2$ on 1024 grid points with the explicit midpoint method, a two-stage Runge--Kutta method, and a time step limited by both the advection and the diffusion stability conditions.

\subsection{Convection-Diffusion Equation}

\begin{figure}[H]
\centering
\includegraphics[width=\textwidth]{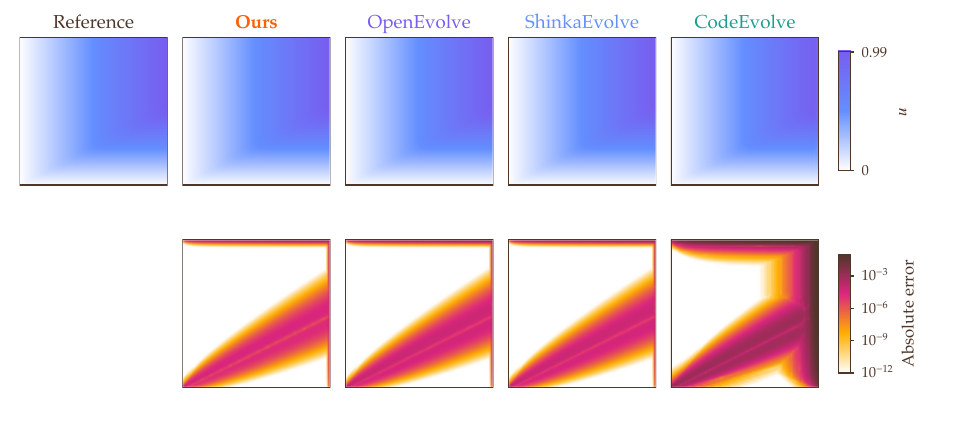}
\caption{Solutions of the convection-diffusion equation for $\varepsilon = 10^{-3}$ (top) and their absolute errors (bottom). The errors concentrate in the boundary layers along $x = 1$ and $y = 1$ and along the characteristic line from the corner $(0, 0)$. The coarser grid of CodeEvolve spreads the error near $x = 1$ over a wider region.}
\label{fig:best-convection-diffusion}
\end{figure}

\autoref{fig:best-convection-diffusion} shows the solution of each framework for $\varepsilon = 10^{-3}$ and its error.

\paragraph{Our method.}
The program uses central finite differences on a $768 \times 768$ grid with a sparse direct solver and bilinear interpolation to the evaluation points.
\paragraph{OpenEvolve.}
The program uses central finite differences on a $512 \times 512$ grid with a sparse direct solver and bilinear interpolation.
\paragraph{ShinkaEvolve.}
The program uses central finite differences on a $512 \times 512$ grid with a sparse direct solver and bilinear interpolation.
\paragraph{CodeEvolve.}
The program uses central finite differences on a $128 \times 128$ grid with a sparse direct solver and bilinear interpolation.

\subsection{Helmholtz Equation}

\begin{figure}[H]
\centering
\includegraphics[width=\textwidth]{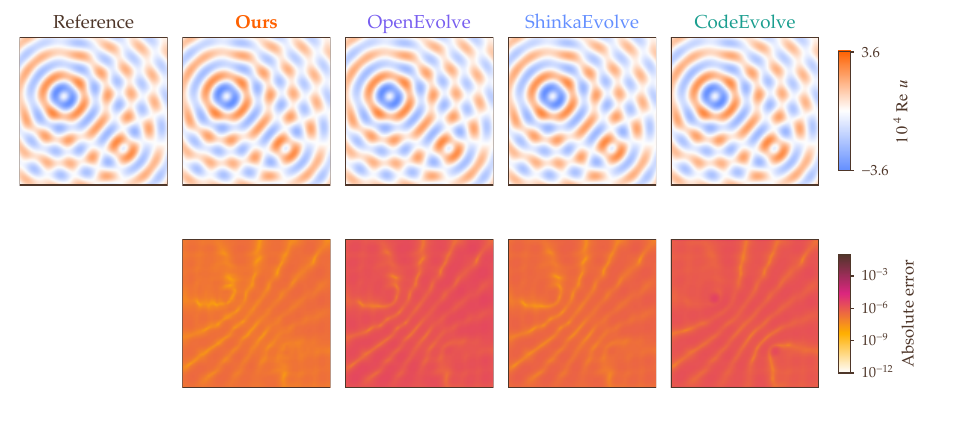}
\caption{Real parts of the solutions of the Helmholtz equation for $k = 40$ (top) and their absolute errors (bottom). The error decreases with the number of grid points per wavelength.}
\label{fig:best-helmholtz}
\end{figure}

\autoref{fig:best-helmholtz} shows the solution of each framework for $k = 40$ and its error.

\paragraph{Our method.}
The program uses second-order finite differences with ghost points for the impedance condition on an $818 \times 818$ grid, a sparse direct solver, and bilinear interpolation.
\paragraph{OpenEvolve.}
The program uses the same discretization on a $301 \times 301$ grid.
\paragraph{ShinkaEvolve.}
The program uses the same discretization on a $601 \times 601$ grid.
\paragraph{CodeEvolve.}
The program uses the same discretization on a $401 \times 401$ grid and smooths the real and imaginary parts of the solution with a Gaussian filter.

\section{Evolution Trajectories}
\label{app:trajectories}

\begin{figure}[H]
\centering
\includegraphics[width=\textwidth]{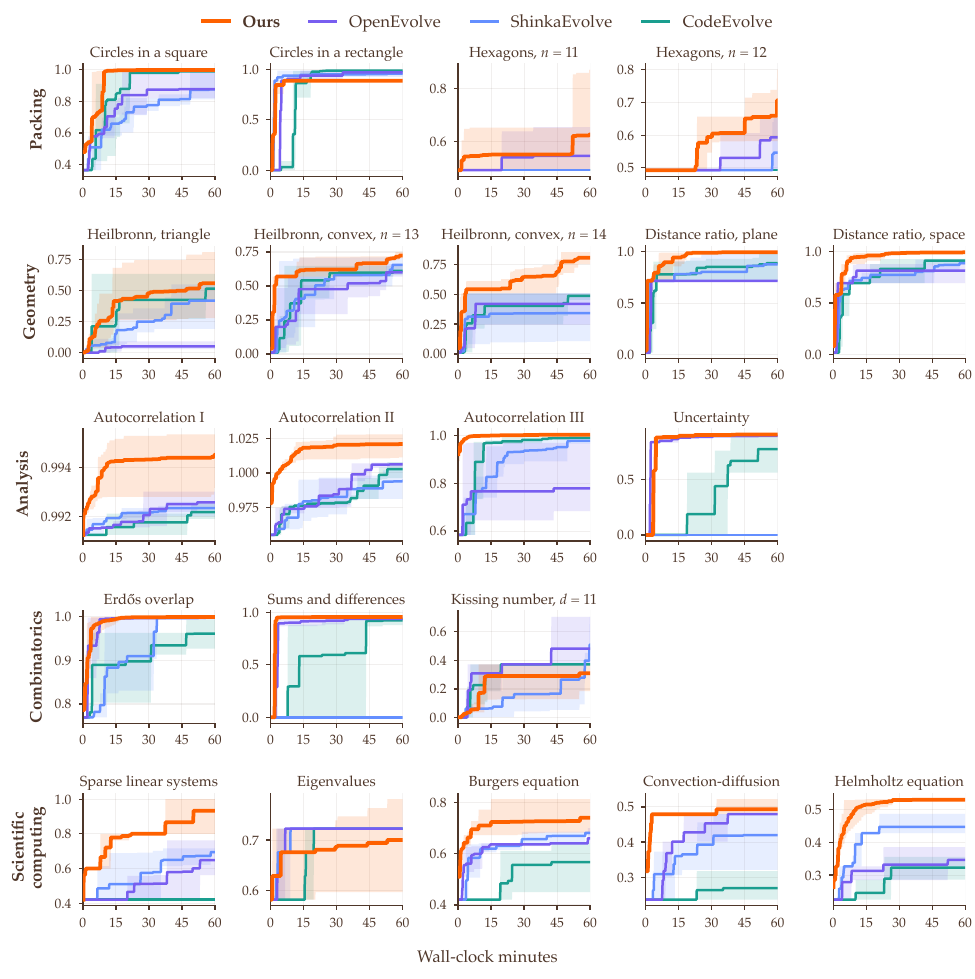}
\caption{Best score so far against wall-clock minutes for each problem with Qwen3 8B. Each line is the best score so far averaged over three runs, and each band spans the lowest and highest run. SwarmEvolve ends the hour with the highest score on 18 of the 21 problems, and its curves rise most steeply in the first few minutes.}
\label{fig:score-over-time-8B}
\end{figure}

\begin{figure}[H]
\centering
\includegraphics[width=\textwidth]{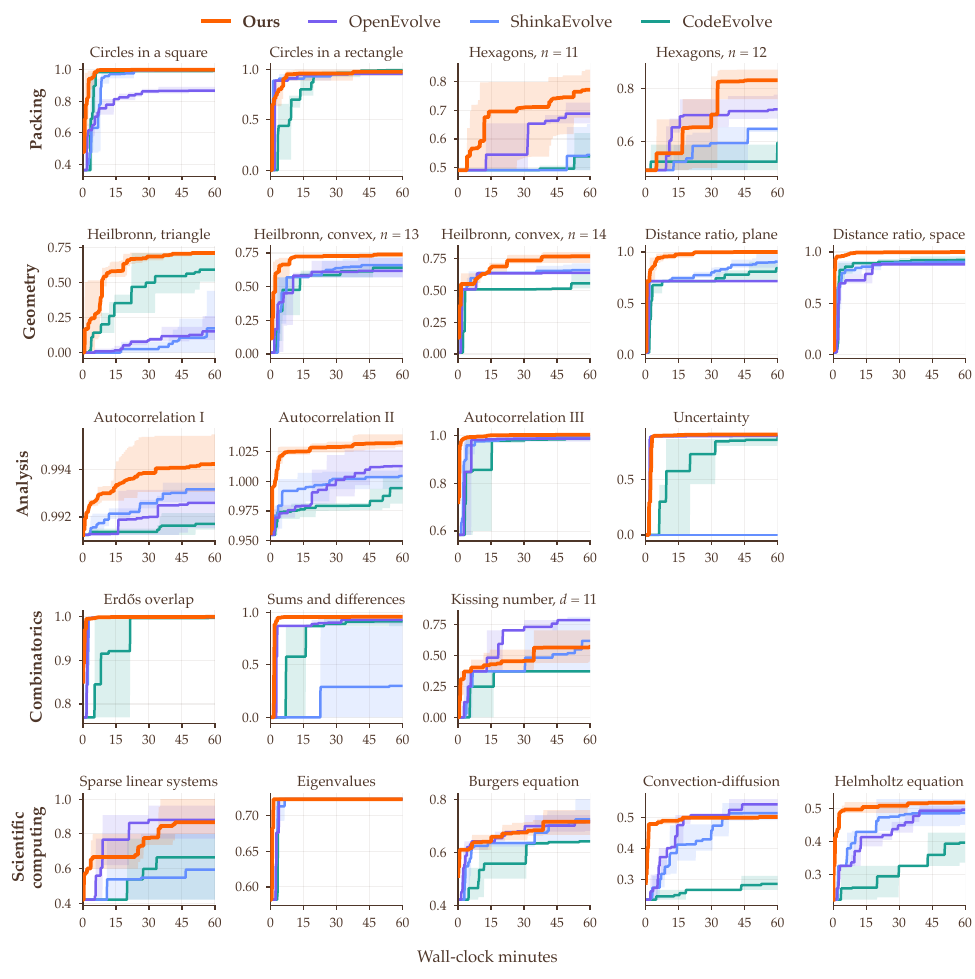}
\caption{Best score so far against wall-clock minutes for each problem with Qwen3 14B. Each line is the best score so far averaged over three runs, and each band spans the lowest and highest run. SwarmEvolve ends the hour with the highest or tied-highest score on 16 of the 21 problems.}
\label{fig:score-over-time-14B}
\end{figure}

\begin{figure}[H]
\centering
\includegraphics[width=\textwidth]{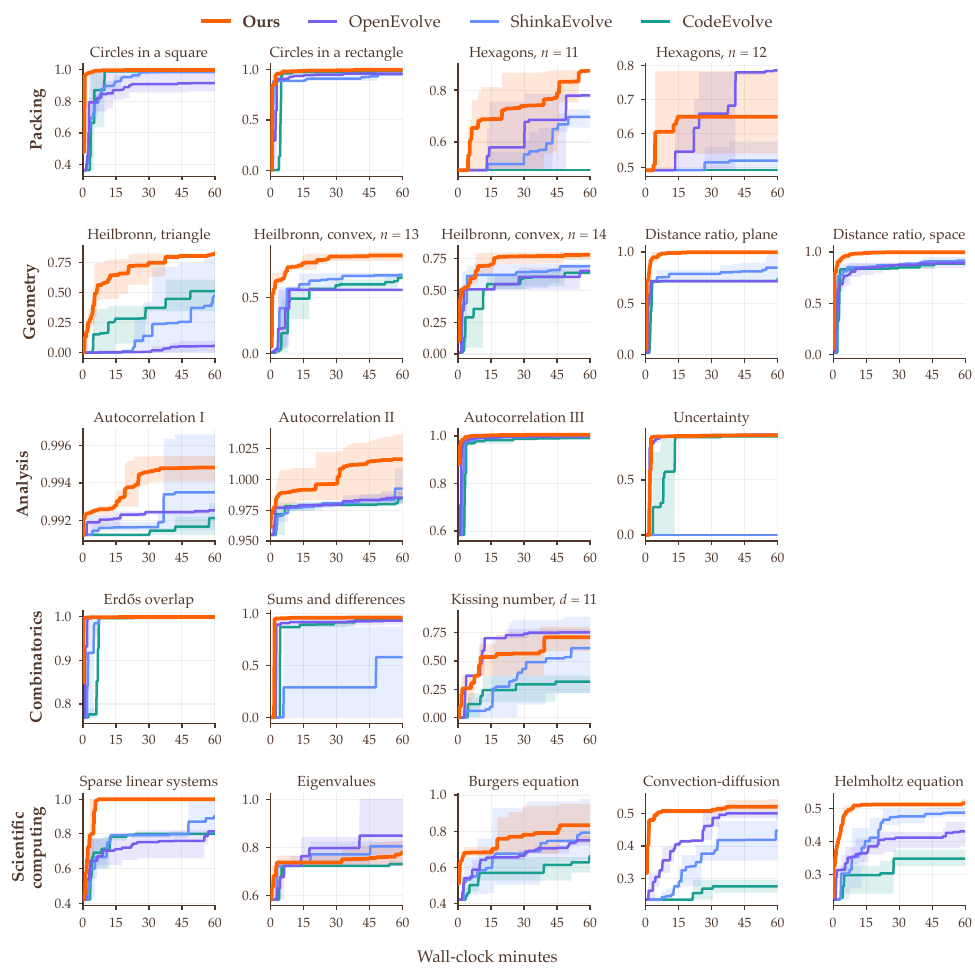}
\caption{Best score so far against wall-clock minutes for each problem with Qwen3 32B-FP8. Each line is the best score so far averaged over three runs, and each band spans the lowest and highest run. SwarmEvolve ends the hour with the highest or tied-highest score on 17 of the 21 problems.}
\label{fig:score-over-time-32B}
\end{figure}

Supplementary Figure~\ref{fig:score-over-time-8B} plots the best score so far against wall-clock minutes for each problem and framework with Qwen3 8B, and Supplementary Figures~\ref{fig:score-over-time-14B} and~\ref{fig:score-over-time-32B} show the same with Qwen3 14B and 32B-FP8. Each line averages the three runs of a framework, and each band spans its lowest and highest run. With every model, the curves of SwarmEvolve rise most steeply in the first minutes, when its rounds explore many focuses at once, and end the hour highest on most problems. Its bands often lie entirely above those of the baselines, so its lead holds for every seed.

\end{document}